\documentclass[10pt,journal,compsoc]{IEEEtran}

\ifCLASSINFOpdf
\else
\fi
\usepackage[export]{adjustbox}
\usepackage{hyperref}
\usepackage{graphicx}
\usepackage{times}
\usepackage{amsmath}
\usepackage{amssymb}
\usepackage{url}
\usepackage{hyperref}
\usepackage{makecell}
\usepackage{ctable}
\usepackage{multirow}
\usepackage{stfloats}
\usepackage{array}
\usepackage{gensymb}
\usepackage{balance}
\usepackage{float}
\usepackage{caption}
\usepackage{subcaption}
\usepackage[normalem]{ulem}

\newcolumntype{Q}{>{\centering\arraybackslash}m{1cm}}
\newcolumntype{O}{>{\centering\arraybackslash}m{0.8cm}}
\newcolumntype{P}{>{\centering\arraybackslash}m{1.4cm}}
\newcolumntype{L}{>{\centering\arraybackslash}m{1.5cm}}
\newcolumntype{S}{>{\raggedright\arraybackslash}m{3.8cm}}
\newcolumntype{K}{>{\raggedright\arraybackslash}m{2.3cm}}
\newcolumntype{A}{>{\centering\arraybackslash}m{1.5cm}}
\newcolumntype{R}{>{\centering\arraybackslash}m{0.65cm}}
\newcolumntype{V}{>{\centering\arraybackslash}m{1.3cm}}

\begin{document}

\title{A Guide to Image and Video based Small Object Detection using Deep Learning : Case Study of Maritime Surveillance}

\author{Aref Miri Rekavandi, \textit{Member, IEEE,} Lian Xu,  Farid Boussaid, Abd-Krim Seghouane, \textit{Senior Member, IEEE,} Stephen Hoefs, and Mohammed Bennamoun, \textit{Senior Member, IEEE,} 
\thanks{Aref Miri Rekavandi, Lian Xu, and Mohammed Bennamoun are  with  the  Department of Computer Science and Software Engineering, The University of Western Australia, 35 Stirling Highway, Crawley, WA, 6009, Australia.}
\thanks{Farid Boussaid is with the Department of Electrical, Electronics and Computer Engineering, The University of Western Australia, 35
Stirling Highway, Crawley, WA, 6009, Australia.}
\thanks{Abd-Krim Seghouane is with the School of Mathematics and Statistics, The University of Melbourne, Melbourne,  Australia.}
\thanks{Stephen Hoefs  is discipline leader of submarine optronics, undersea combat systems, and undersea command and control
maritime division, Defence Science and Technology Group, Australia. (emails: aref.mirirekavandi@uwa.edu.au, lian.xu@uwa.edu.au, farid.boussaid@uwa.edu.au, abd-krim.seghouane@unimelb.edu.au, stephen.hoefs@defence.gov.au,  and mohammed.bennamoun@uwa.edu.au}}

\markboth{}%
{Miri \MakeLowercase{\textit{et al.}}: A Guide to Image and Video based Small Object Detection using Deep Learning : Case Study of Maritime Surveillance}

\date{}

\IEEEtitleabstractindextext{%

\begin{abstract}
Small object detection (SOD) in optical images and videos is a challenging problem that even state-of-the-art generic object detection methods fail to accurately localize and identify such objects. Typically, small objects appear in real-world due to large camera-object distance. Because small objects occupy only a small area in the input image (\textit{e.g.}, less than 10\%), the information extracted from such a small area is not always rich enough to support decision making. Multidisciplinary strategies are being developed by researchers working at the interface of deep learning and computer vision to enhance the performance of SOD deep learning based methods. In this paper, we provide a comprehensive review of over 160 research papers published  between 2017 and 2022 in order to survey this growing subject. This paper summarizes the existing literature and provide a taxonomy that illustrates the broad picture of current research. We investigate how to improve the performance of small object detection in maritime environments, where increasing performance is critical. By establishing a connection between generic and maritime SOD research, future directions have been  identified. In addition, the popular datasets that have been used for SOD for generic and maritime applications are discussed, and also well-known evaluation metrics for the state-of-the-art methods on some of the datasets are provided.
\end{abstract}

}

\maketitle

\IEEEpeerreviewmaketitle

\section{Introduction}
\IEEEPARstart{O}{bject} detection is at the heart of many computer vision applications and has grown in importance over the last decade. It plays a crucial role in modern computer vision tasks such as autonomous driving \cite{chen2016monocular,wang2020pillar}, pedestrian identification \cite{liu2019high,lan2018pedestrian}, image captioning \cite{herdade2019image,iwamura2021image}, object tracking \cite{yin2021center,lee2021cnn}, ship detection \cite{liu2017rotated,rekavandi2021robust} face recognition \cite{yan2022age, li2022enhanced}, traffic control \cite{khan2022machine,ge2022vehicle}, animal detection \cite{berg2022weakly,xue2022small}, action recognition \cite{kanimozhi2022key,patil2022survey}, environment surveillance \cite{jha2021real,kumar2010addressing}, video checking in sports \cite{roros2022maskgru,li2022automatic}, and many others. Object detection methods have become increasingly popular with the advances in deep learning and GPU power that allow Deep Neural Nets (DNNs) to be trained faster and more efficiently in recent years. Object detection methods are classified into two-stage and single stage methods. A few notable two-stage methods include Region-Based CNN (R-CNN) \cite{girshick2014rich}, Spatial Pyramid Pooling Network (SPP-Net) \cite{he2015spatial},  Fast R-CNN \cite{girshick2015fast}, Faster R-CNN \cite{ren2015faster}, Region-Based Fully Convolutional Networks (R-FCN) \cite{dai2016r}, Mask R-CNN \cite{he2017mask}, Feature Pyramid Networks  (FPN) \cite{lin2017feature}, cascade R-CNN \cite{cai2018cascade}, and  Libra R-CNN \cite{pang2019libra}. These methods identify the regions in an image that are most likely to contain objects, then features are extracted to classify the objects, followed by a fine-tuning step to accurately localize the bounding boxes surrounding the objects. Some anchor-free (anchor defines a predefined set of bounding boxes with a particular height and width) detectors such as RepPoints \cite{yang2019reppoints} can also be viewed as two-stage methods. On the other hand, single-stage methods treat the object detection task as a regression problem and estimate the parameters of the bounding boxes and the probability that these boxes contain the target objects. This category of methods includes You Only Look Once (YOLO) and its variants \cite{redmon2016you,redmon2017yolo9000,redmon2018yolov3,bochkovskiy2020yolov4,jocher2020yolov5}, Single Shot multibox Detector (SSD) \cite{liu2016ssd}, RetinaNet \cite{lin2017focal}, Multi-Scale Deep Feature
Learning Network (MDFN) \cite{ma2020mdfn} and anchor-free object detection methods such as CornerNet \cite{law2018cornernet}, CenterNet \cite{duan2019centernet}, FCOS \cite{tian2019fcos}.

Although the above mentioned object detection techniques have undoubtedly grown due to the availability of large datasets, \textit{e.g.}, ImageNet \cite{russakovsky2015imagenet}, PASCAL VOC \cite{everingham2010pascal} and MS COCO \cite{lin2014microsoft}, most of these deep learning based techniques fail to accurately localize and identify small objects. The main reason for their poor performance to deal with small objects is due to the loss of the geometrical information in the last layers of their networks and their large receptive fields. Solely the semantic information recovered from the last layers of deep neural networks is indeed useful for larger objects classification, but cannot help with the localization of small objects. Max pooling or large steps toward down sampling are responsible for the large receptive fields of the convolutioanl layers, \textit{e.g.}, $\times8$ and $\times 32$ in SSD and YOLO. As a result, the last layers of deep networks have a small number of nodes whose values reflect the small objects in the input image, which is not desirable for SOD.\\
The applications of small object detection (SOD) are but not limited to pedestrian detection \cite{song2018small,wu2020self}, medical image analysis \cite{xing2016robust,rashidi2020optimal}, industrial product quality assessment \cite{abedini2017defect}, face recognition in surveillance cameras \cite{cho2018face}, sign detection in autonomous driving \cite{li2022trafficss}, ship detection in remotely sensed images \cite{liu2017rotated} and others. In spite of the extensive potential use of SOD methods in the maritime surveillance, unlike the other applications, this area has not been explored as much as it truly deserves. This may be the result of the paucity of publicly available datasets for maritime environment, as compared to datasets for other applications.

\begin{figure*}
	
	\begin{tabular}{@{\hskip1pt}c@{\hskip3pt} c @{\hskip3pt} c @{\hskip3pt}c@{\hskip3pt}}
		 Airplane &\includegraphics[width=0.28\linewidth]{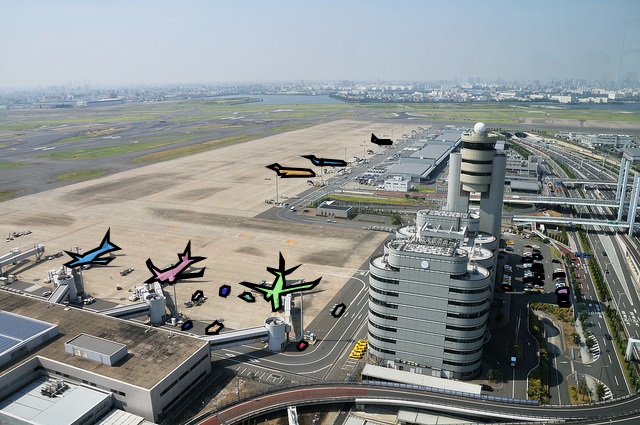} &\includegraphics[width=0.29\linewidth]{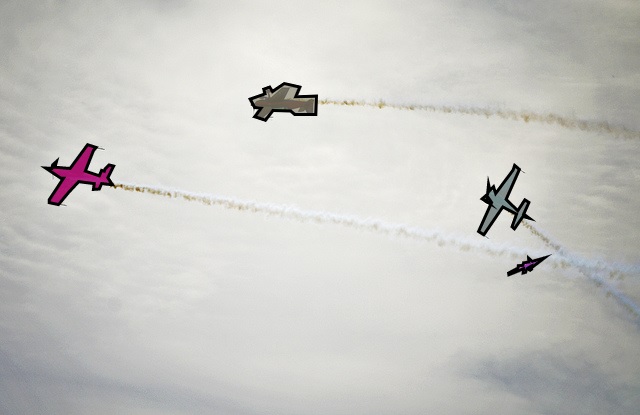}&\includegraphics[width=0.28\linewidth]{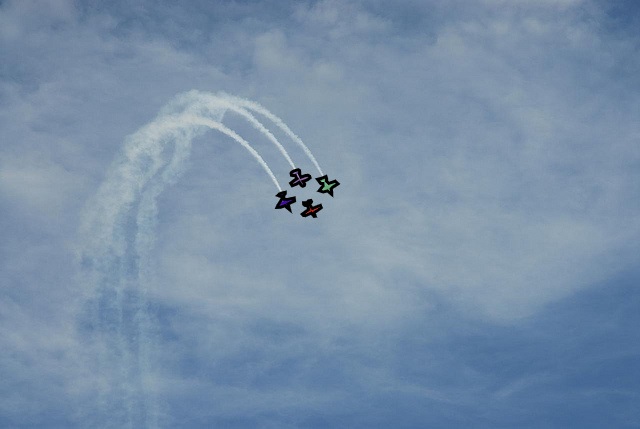}\\
		\hline
				Traffic Light &\includegraphics[width=0.28\linewidth]{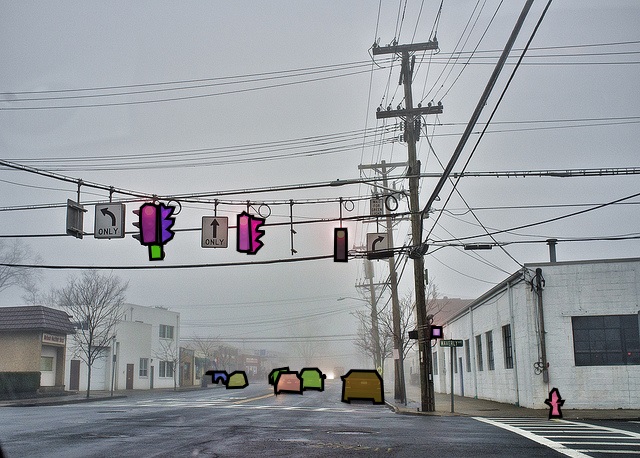} &\includegraphics[width=0.28\linewidth]{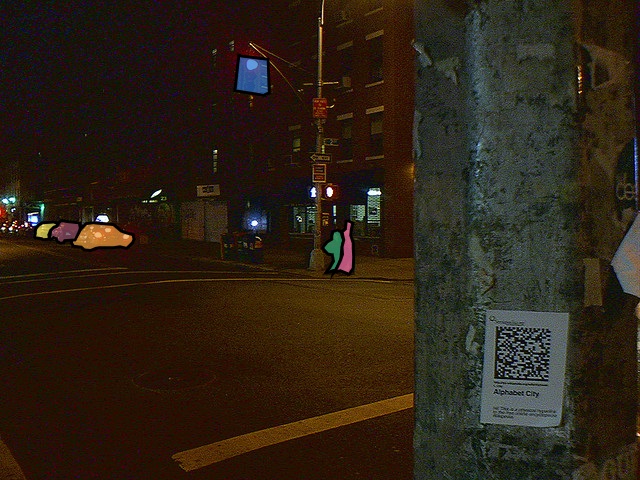}&\includegraphics[width=0.28\linewidth]{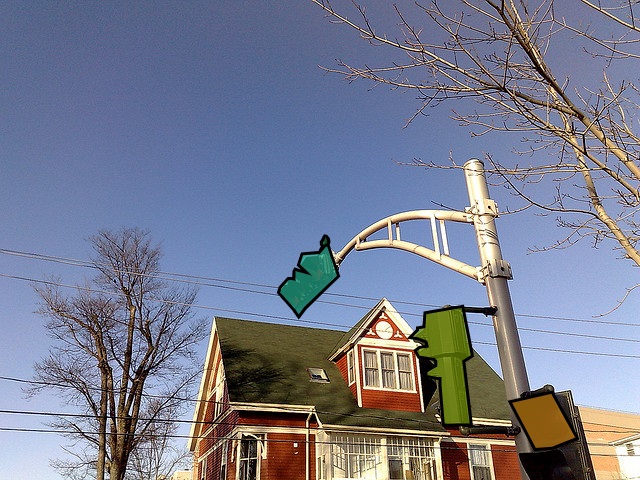}\\
		\hline
						Person &\includegraphics[width=0.28\linewidth]{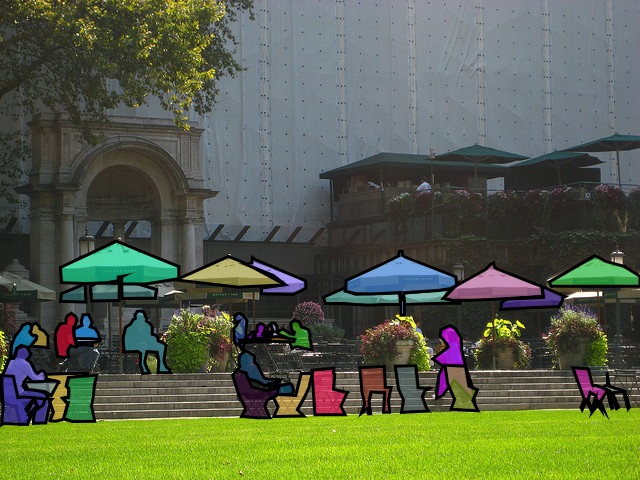} & \includegraphics[width=0.28\linewidth]{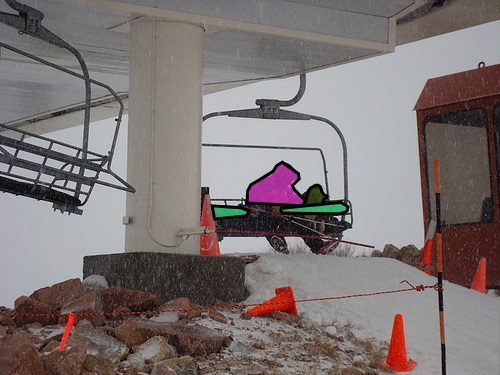}&\includegraphics[width=0.28\linewidth]{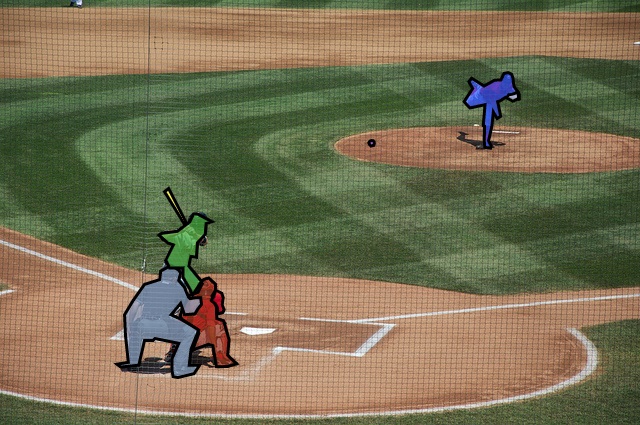}\\
							\hline
													Boat &\includegraphics[width=0.28\linewidth]{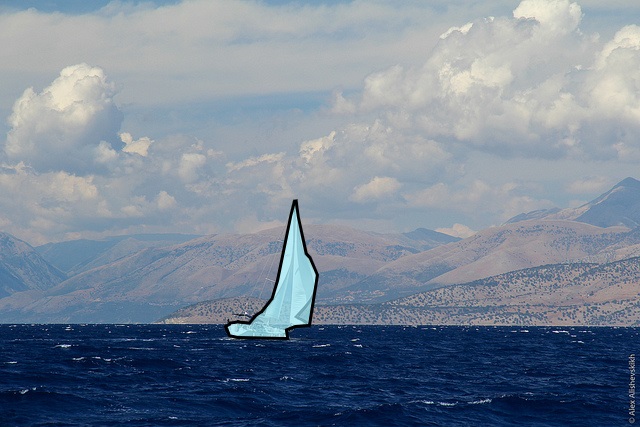} & \includegraphics[width=0.28\linewidth]{Figures/boat2.jpg}&\includegraphics[width=0.28\linewidth]{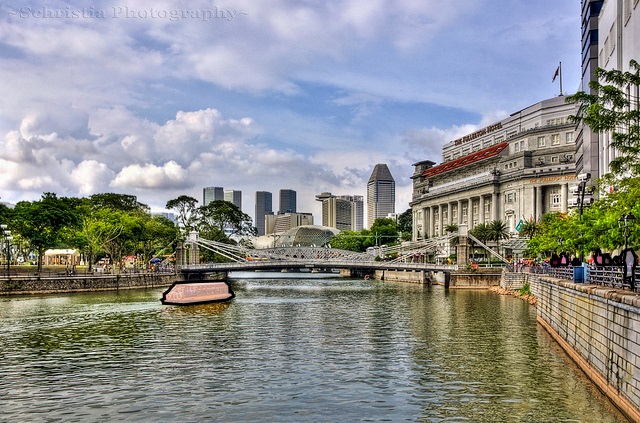}\\
							\hline
	\end{tabular}	
	\caption{Examples of small objects. Source: MS COCO dataset \cite{lin2014microsoft}. By definition, small objects refer to the objects smaller than $32 \times 32$ pixels or objects which cover less than only 10\% of the image. }
	\label{fig2}
\end{figure*}
Approximately $70\%$ of the planet is covered by water, so most of the global trade and transportation of goods takes place by sea \cite{international2011international}. This requires accurate monitoring of the environment for rescue missions, and to avoid collisions, pollution from oil leaks, illicit cargos, illegal smuggling, fisheries dumping of pollutants, and the crossing of borders by unidentified vessels. In spite of the fact that an Automatic Identification System (AIS) can be used to monitor vessels, many small and even medium-sized vessels lack such technology, or intentionally switch it off when they conduct illegal activities. Therefore, the development of a wide range automatic system that is capable of detecting and identifying small boats is vital. Synthetic Aperture Radar (SAR) technology has been the leading technology since the 1990s, providing an all-time performance and a strong signal reflection response from normal large vessels. However, the relatively weak reflected signal from small or medium-sized targets with small radar cross-sections makes it difficult to recognize targets due to the observed speckle multiplicative noise, resulting in a high number of false positives. Furthermore, SAR cannot provide global range monitoring because of its limited spatio-temporal coverage. This opens up a wide range of research opportunities in maritime environments, including the detection of objects based on images and videos.

A variety of definitions have been reported for ``small object" in the literature, but most studies define a small object as one that is  smaller than $32 \times 32$ pixels.  In high resolution images, a small object is one that covers less than 10\% of the image \cite{lin2014microsoft}. This definition means that the object of interest does not provide much information in terms of colour, shape, texture, or any other type of visually discernible information, making the task of SOD particularly challenging. There are mainly two reasons why small objects appear in images and videos. \textbf{First}, the object appears small by virtue of its size, \textit{e.g.}, a bird relative to a tree, a tennis ball relative to a tennis court, or a mobile phone relative to an indoor space, and so forth.  \textbf{Second}, a large object-camera distance can also lead to the object looking small, in which case the object's real size is irrelevant. Even a ship can appear small and occupy only a few pixels in a satellite image. Fig. \ref{fig2} shows examples of small objects.
\begin{figure*}[!t]
 {
 \includegraphics[scale=.8]{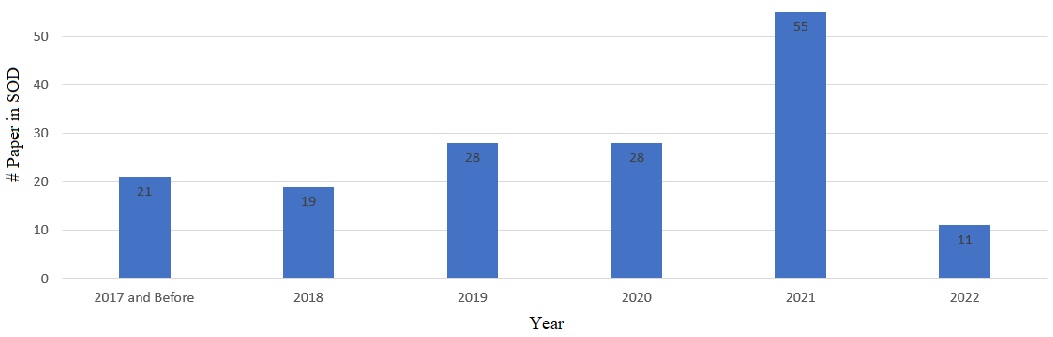}}
 \centering
 \caption{Distribution of reviewed SOD-based papers in this study over time.}
   \label{time}
 \end{figure*}

The task of small object detection is typically performed through a variety of computer vision techniques, such as semantic segmentation, foreground background (FB) separation, anomaly detection, regression, and finally classification. Many data modalities have also been explored in the context of SOD in the literature, including AIS data, satellite-based SAR and multi-spectral data, airborne SAR, multi-spectral data from Unmanned Aerial Vehicles (UAVs), on board (ship based, unmanned surface vessels, \textit{etc.}) visual (RGB video and image), InfraRed (IR) and Near InfraRed (NIR) data, and finally shore based which includes visual data (RGB video and image), \textit{etc.} Often these modalities differ in terms of their spatial and temporal resolution, cost of acquiring data, delay, robustness, range of coverage, \textit{etc.} \cite{kanjir2018vessel}. Spaceborne data (satellite), for example, can be accessed remotely. Satellites positioned in geostationary orbits may also capture images of the surface of the earth while maintaining the same footprint. Data volume generated by this technology is quite large, and it is often not suitable for continuous monitoring \cite{kanjir2018vessel}. Furthermore, spaceborne optical images are affected by bad weather (clouds covering objects of interest), while radar data has a low resolution. Infrared imaging is particularly well-suited for night-time monitoring. However, it becomes saturated during the daytime and it does not provide colour information. Optical imaging on the other hand, provides rich colour information, real-time operation, adequate spatial resolution, and is relatively inexpensive. In particular, spaceborne optical sensors are growing in number and are becoming increasingly popular because of their excellent spatial coverage. For this reason, this survey paper will focus on images or videos acquired by optical cameras, from space, air, in-shore and off-shore.
 
\begin{figure*}[!h]
 {
 \includegraphics[scale=.50]{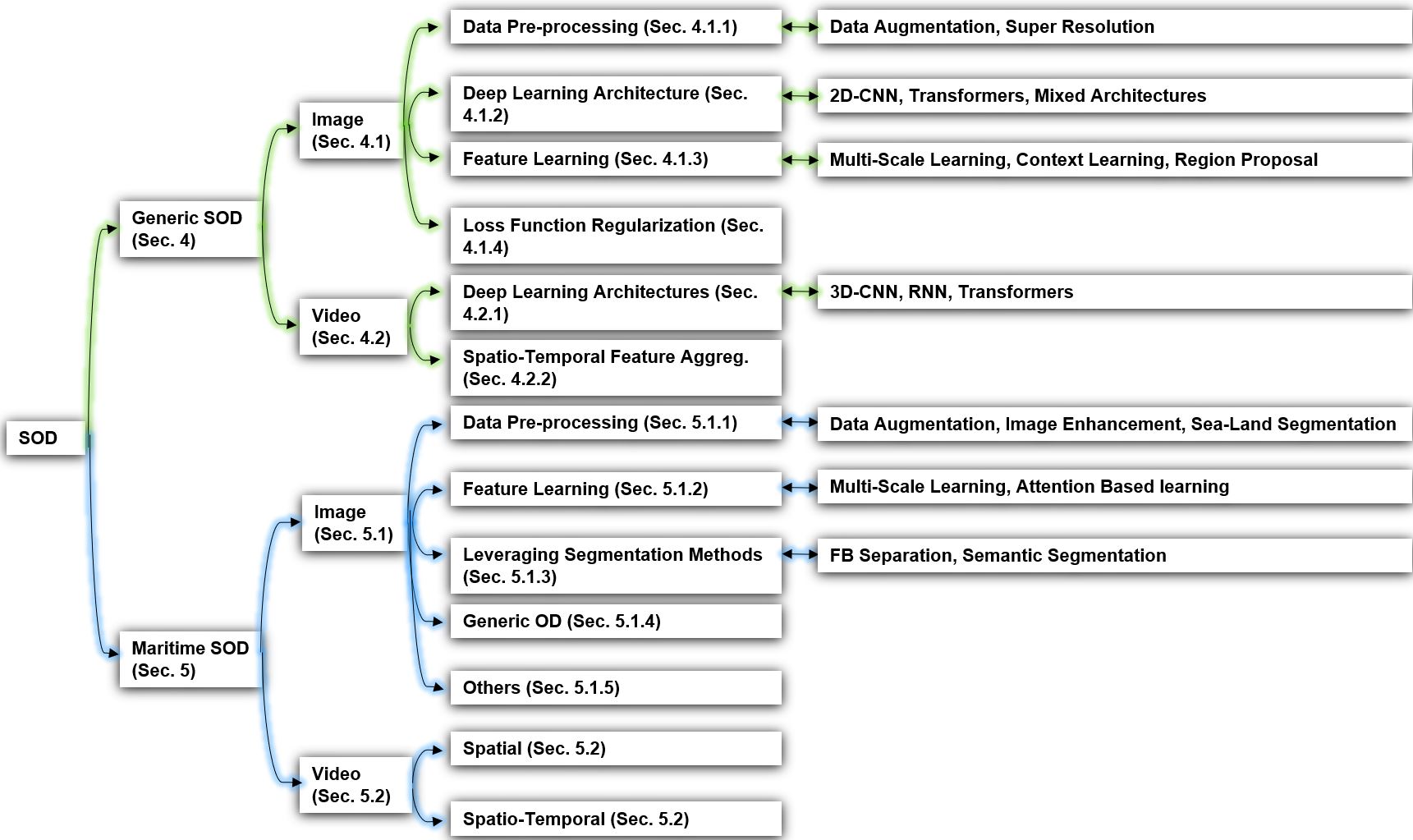}}
 \centering
 \caption{Taxonomy of small object detection in images and videos. This taxonomy follows the paper's organization and divides the literature into generic and maritime specific methods. }
   \label{tax}
 \end{figure*}

Specifically, this paper will review the field of small object detection using deep learning, with a case study covering maritime applications. Our literature survey was conducted by searching for keywords such as ``small object detection", ``small target detection", ``tiny object detection", and ``ship detection" in title. Checking the corresponding references of individual papers on Google Scholar also yielded a comprehensive list of studies. We limited the scope of this survey to deep learning based methods. Our survey paper reviewed more than 160 papers, most of which were published after 2017 (Fig. \ref{time}), when deep learning methods began to show promising results for object detection. Small object detection is a relatively new field, so this survey provides an overview of the current state-of-the-art (SOTA) and may also serve as a guide for upcoming research. In summary, the contributions of this survey paper are as follows:
\begin{table*}[t!]%
\caption{A list of the recently published surveys on maritime and generic SOD.}

  \centering
    \begin{tabular}{|p{4cm}|l|p{2.5cm}|p{1cm}|p{1.5cm}|p{2.5cm}|p{2.5cm}|}
    \specialrule{.15em}{.0em}{.0em} 
    \textbf{Survey Title} & \textbf{Year} & \textbf{Publisher} & \textbf{Category} &  \textbf{Image/Video}& \textbf{Limitations} &\textbf{Strengths} \\\specialrule{.15em}{.0em}{.0em} 
        Video processing from electro-optical sensors for object detection and tracking in a maritime environment: a survey \href{https://ieeexplore.ieee.org/abstract/document/7812788?casa_token=-K5wsS86f3kAAAAA:ejknrtfXZcEhNL6h8FG2INcDmRENoyICBlriKHgIEPm2HXi3_DF1P9fMz7hu1_XYVXONswc9}{\cite{prasad2017video}} &2017&IEEE Transactions on Intelligent Transportation Systems&Maritime&Video&It just covers the classical methods not the DNNs& Both Visible and NIR parts of the spectrum \\
    \hline
    Vessel detection and classification from spaceborne optical images: A literature survey \href{https://www.sciencedirect.com/science/article/pii/S0034425717306193}{\cite{kanjir2018vessel}} & 2018& Remote Sensing of Environment& Maritime & Image &  This survey is up to 2017, does not contain deep learning based methods, constrained to spaceborne images &  Covers all the classical approaches multiple data modalities in details \\
    
    \hline
    Recent advances in small object detection based on deep learning: A review \href{https://www.sciencedirect.com/science/article/pii/S0262885620300421?casa_token=yE03MSmyqPcAAAAA:DG1Du6EMUZgmjp8bn_wTExJhrPIOpW1NPbb5SPUZ8zrFRcLqfh8ABShweDPuHL1_fg1AF-13bQ}{\cite{tong2020recent}} & 2020 & Image and Vision Computing & Generic & Image & Their taxonomy is very general, does not cover maritime environment, does not cover video & It gives a great list of the works up to 2020 for deep learning based methods\\
    \hline
   
    Ship detection and classification from optical remote sensing images: A survey \href{https://www.sciencedirect.com/science/article/pii/S1000936120304544}{\cite{bo2021ship}}  & 2021 & Chinese Journal of Aeronautics & Maritime & Image &This survey is up to 2020, constrained to remote sensing images, Not detailed for DNNs&  To an extent at time of publication is up to date and includes some DNN based methods\\
     \hline
    Survey on Deep Learning-Based Marine Object Detection \href{https://www.hindawi.com/journals/jat/2021/5808206/}{\cite{zhang2021survey}} &2021&Journal of Advanced Transportation&Maritime&Image \& Video & It does not categorize the studies based on their adopted appraches, does not introduce available datasets, does not emphasize on SOD & A recent review which to an extent includes deep learning methods for maritime up to 2021\\
    \hline
    Survey of Video Based Small Target Detection \href{chrome-extension://gphandlahdpffmccakmbngmbjnjiiahp/http://www.joig.net/uploadfile/2021/1124/20211124052219501.pdf}{\cite{liu2021survey}} & 2021 & Journal of Image and Graphics & Generic & Video & It focuses mostly on spatial methods, instead of spatio temporal, datasets are not comprehensive & Recent video based detection survey for SOD, addresses studies up to 2021\\
    \hline
    A survey of the four pillars for small object detection: Multiscale representation, contextual information, super-resolution, and region proposal \href{https://ieeexplore.ieee.org/abstract/document/9143165?casa_token=sIWEf-d7_4gAAAAA:3EH6ilJToUVxuDiZuFlrdw4jTWdCEMQf3jIUvvv6cdoGn976hGzo55_aintUlqcV9kZ-YygT}{\cite{chen2020survey}}  & 2022 & IEEE Transactions on systems, man, and cybernetics: systems & Generic & Image & The aerial perspective is not included, limited datasets, subsection of the current manuscript.& Divides the prior works into four categories that are somehow related to popular object detection frameworks\\
    \hline
    A Guide to Image and Video based Small Object
    Detection using Deep Learning : Maritime
    Surveillance Case Study (\textbf{Ours})& 2022 & ArXiv & Generic \& Maritime & Image \& Video & Limited to optical images and only DNN based techniques  &  We cover state-of-the-art methods in DNNs including transformers, We cover both image and video, we list all the available datasets in detail, we suggest very diverse future research directions \\
    \hline

    \end{tabular}%
    \label{tab:idl}
\end{table*}%
\begin{itemize}
    \item First, we review generic small object detection methods. This is the first review that explores both image and video modalities for small object detection using deep learning frameworks, including both CNNs and transformers (transformers have not previously been covered previously in any survey). Our careful review of the literature has allowed us to identify research gaps and suggest potential research directions.
    \item Our study has identified object detection in maritime environments as an important and challenging task, and in addition to generic SOD, we also present a systematic review of SOD in maritime environments.
    \item By comparing and establishing links between the literature of generic and maritime SOD, possible research directions are highlighted for both domains.
    \item There is a limited number of datasets available, and we believe that is the main hurdle for researchers who do not work in this field of research. Therefore, in order to allow future research to be explored more effectively, we have compiled the most relevant and comprehensive datasets (50 datasets) specific to SOD.
    \item Finally, the limitations of existing works as well as possible future directions, and potential tools that could be useful for SOD have been identified.
\end{itemize}
Review papers for SOD are listed in Table \ref{tab:idl}. Our paper differs from existing surveys in that we consider both image and video modalities, look at each component of learning pipeline from the input to the output, establish and discuss the link between maritime and generic SOD to identify research gaps, and introduce the recent deep learning methods that have been proposed up to May 2022. Fig. \ref{tax} shows a taxonomy of small object detection methods, where the works are divided into categories according to their methodology, domains, and applications.\\ 
The remainder of the paper is organized as follows: The challenges of SOD are discussed in Section \ref{challenge}. Section \ref{backgroundsec}, summarizes existing single- and double-stage detectors and the well-used backbones in the context of SOD. In Sections \ref{generic} and \ref{maritime}, we examine generic and maritime SOD methods. We provide evaluation metrics and datasets in Section \ref{dataset} and compare and discuss methods and potential reserach gaps in Section \ref{discussion}. Finally, the paper concludes in Section \ref{conclusion}.  
\section{Challenges in SOD}\label{challenge}
Let's explore some of the potential challenges that potential SOD users may encounter before we delve into the technical content and methodologies. While some of these challenges are common across generic and maritime domains, others are specific to the maritime environment. Listed below are the most common challenges of SOD that fall under maritime specific and generic SOD. Here are some challenges associated with generic SOD:
\begin{itemize}
    \item As a result of the small number of pixels representing each object, SOD loses geometrical information in the deeper layers of the network, resulting in false object detection. 
    \item Small objects are usually occluded by larger objects, and their extracted features behave like clutter because of their relatively weak feature values.
    \item Object detection evaluation metrics that are commonly used are not appropriate for small objects. These metrics can become quite sensitive when the bounding boxes are small, leading to the underestimation of methods or even incorrect solutions.
    \item Compared to regular-size object detection, very few small object datasets have been released to date.
    \item To annotate the ground truth frames between the ground truth human annotated frames in video object detection, most commonly used softwares use interpolation to draw the bounding boxes (\textit{e.g.}, they annotate the 1st and 10th frames, assume linear motion, and use linear interpolation to annotate the frames in between). This is not an issue for large object detection, however it may produce very noisy ground truth labels for SOD. SOD methods should therefore be robust to such deviations.
\end{itemize}
Challenges associated with the detection of small objects in maritime environments include:
\begin{itemize}
    \item The reflection of light from the water and waves can cause rapid changes in illumination in video frames.
    \item The dynamic nature of maritime environments and  challenging weather conditions significantly reduce the range of sight and make the images blurry or hazy. As a result, such environmental factors can adversely affect detection performance, especially when using passive remote sensing imaging to detect ships.
    \item Most of the maritime datasets are aerial. Consequently, depending on the viewing angle and relative position of the target, the object may appear distorted in the image or can appear at different scales, structures and shapes which makes the detection more challenging.
    \item A ship dataset can show greater intra-class variation than inter-class variation, increasing the complexity of maritime SOD.
    \item When aerial data is acquired, the camera's perspective towards the object can rapidly change between frames. A highly dynamic scenario like this can result in the object being missed in SOD over many frames.
    \item Especially for cameras installed on ships, the image data shows jitter at high frequencies and a shift in the field of view at low frequencies due to irregular jittering, hull swaying, and hull heaving \cite{qiao2021marine}.
\end{itemize}
\begin{figure*}[]
	\centering
	\scalebox{0.9}{\begin{tabular}{c|c }
		 \includegraphics[width=0.5\linewidth]{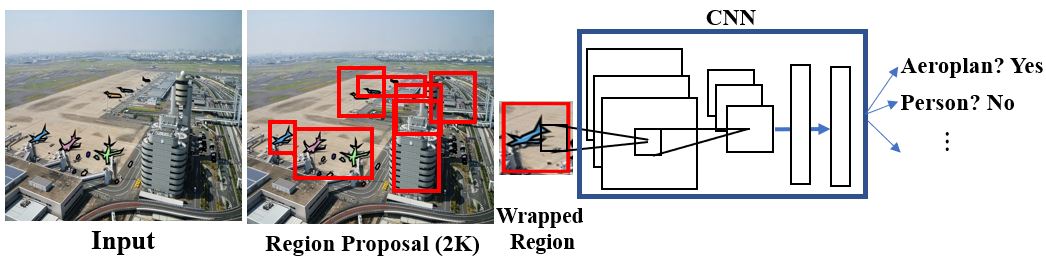} &\includegraphics[width=0.5\linewidth]{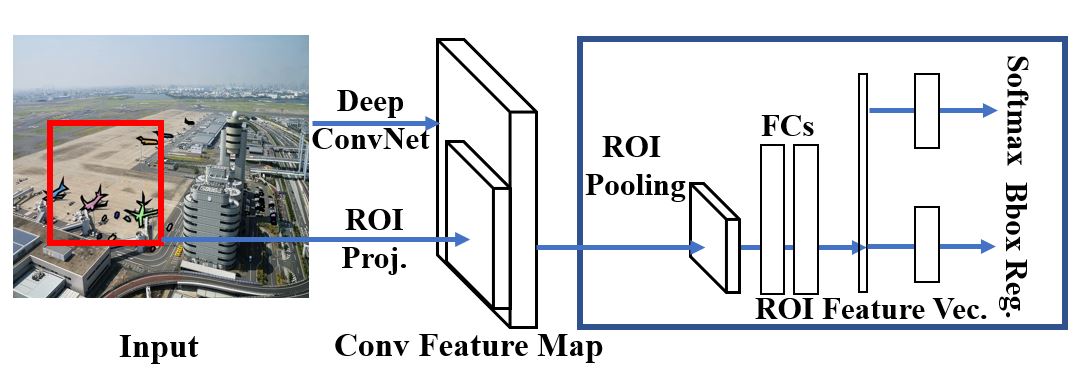}\\
		 (a)&(b)\\
		 \hline
		 \includegraphics[width=0.5\linewidth]{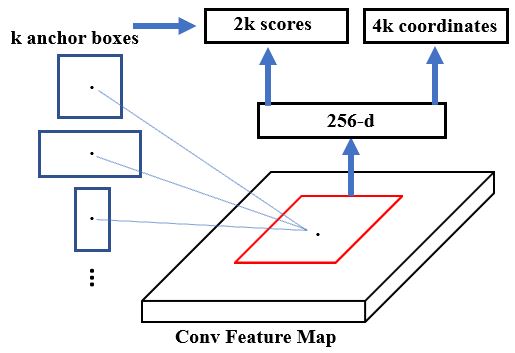} &	\includegraphics[width=0.5\linewidth]{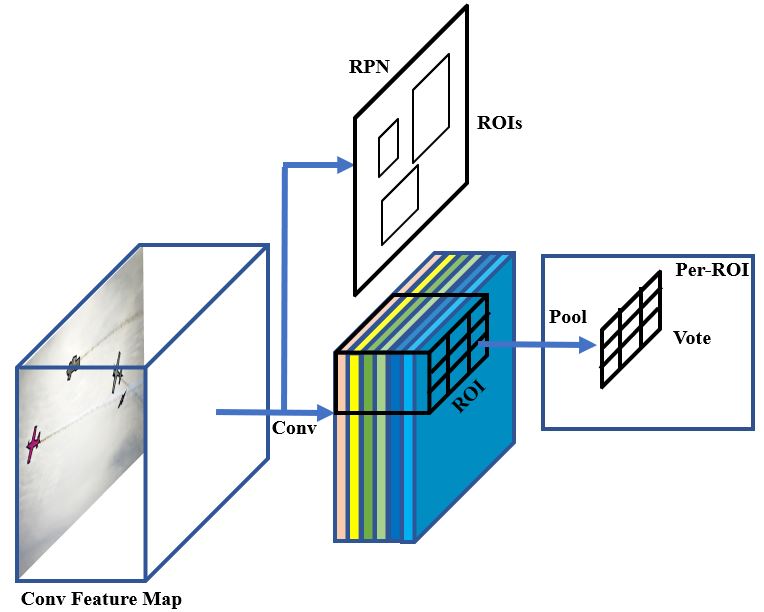}\\
		  (c)&(d)\\
		 \hline
		 \includegraphics[width=0.5\linewidth]{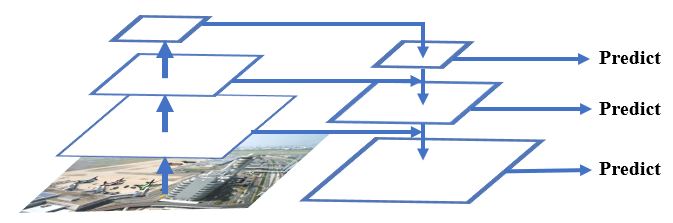} & \includegraphics[width=0.45\linewidth]{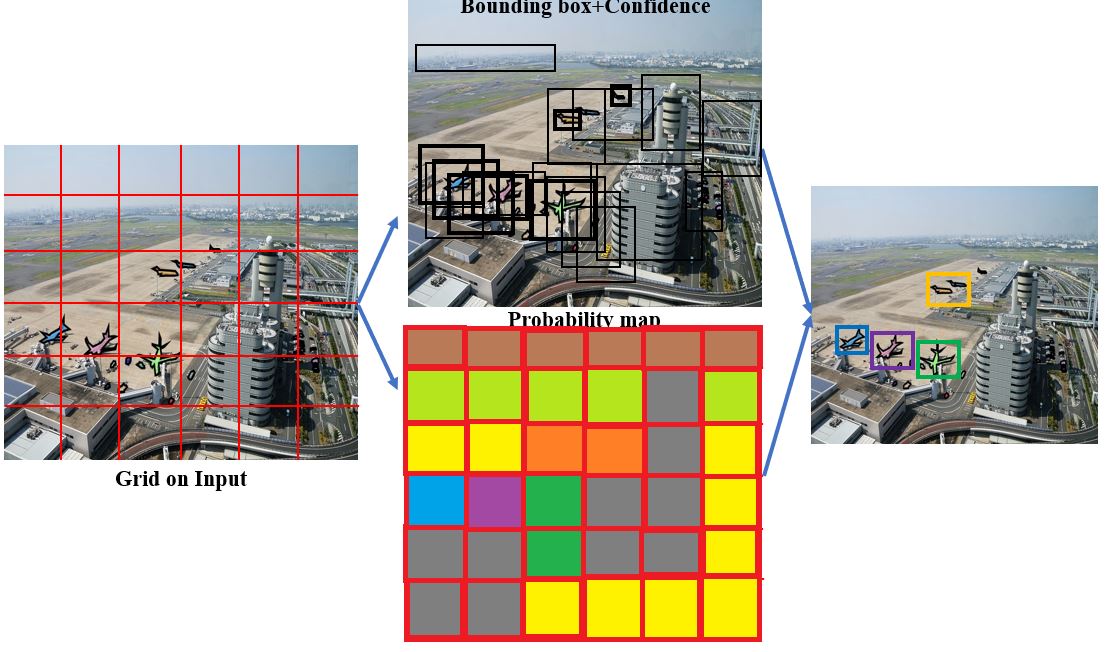}\\
		  (e)&(f)\\
		\hline
		\includegraphics[width=0.5\linewidth]{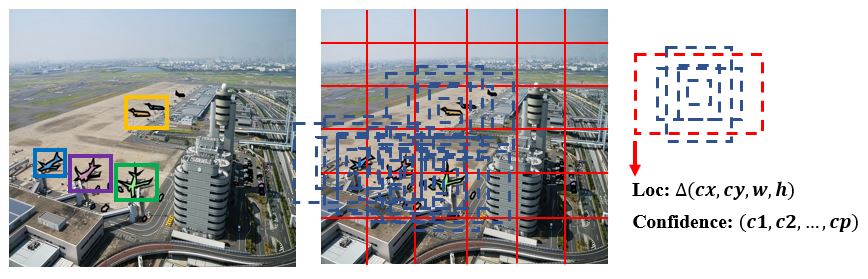} & \includegraphics[width=0.5\linewidth]{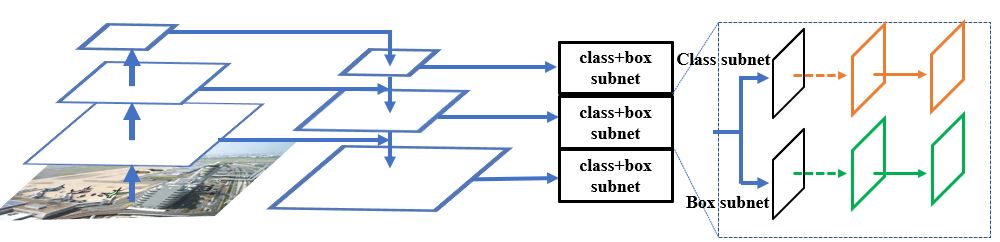}\\
		 (g)&(h)\\
	\end{tabular}}	
	\caption{Popular two-stage and single-stage object detection architectures. Figure adopted from (a) R-CNN \cite{girshick2014rich}, (b) Fast R-CNN \cite{girshick2015fast}, (c) RPN in Faster R-CNN \cite{ren2015faster}, (d) R-FCN \cite{dai2016r}, (e) FPN \cite{lin2017feature}, (f) YOLO \cite{redmon2016you},  (g) SSD \cite{liu2016ssd}, (h) RetinaNet \cite{lin2017focal}.}
	\label{background}
\end{figure*}

\section{Background}\label{backgroundsec}
To ensure completeness, this section provides a brief overview of the most important object detection frameworks that have been used in the SOD literature, including their underlying principles and backbones.\\
\textbf{Regional Based Detectors:} Also known as two-stage detectors, they typically involve the following three main steps: (\textbf{i}) region proposal, (\textbf{ii}) feature extraction, and (\textbf{iii}) classification. The first version of this framework was the Region-Based CNN (R-CNN) \cite{girshick2014rich}, whose pipeline is shown in Fig. \ref{background}(a). R-CNN takes the input image and extracts approximately $2K$ region proposals of different scales using selective search \cite{uijlings2013selective}. In a second step, a CNN is used for feature extraction through five convolutional layers with two fully connected layers (4096-dimensional features), and then SVMs are used for classification. The R-CNN algorithm is relatively slow (two stages) and needs to pass each region individually without sharing computation. In addition, it is trained in multiple stages. R-CNN's first issue is fixed by SPP-Net, which shares computation \cite{he2015spatial}. SPP-Net extracts convolutional feature maps from the entire image and features are extracted from the shared feature maps to classify the objects in region proposals. In this way, the process becomes faster, and the runtime at the test stage is also reduced. In \cite{girshick2015fast}, an extension of R-CNN dubbed Fast R-CNN was proposed to increase the runtime speed of R-CNN and SPP-Net, using a multi-task loss function for learning in a single stage. On the deep VGG16 network, fast R-CNN improves training time by $9\times$ and test time by $213\times$ over regular R-CNN. Fast R-CNN jointly classifies and localizes bounding boxes (Fig. \ref{background}(b)). Faster R-CNN \cite{ren2015faster} was introduced to improve the bottleneck of the two-stage framework, which is the first step of the pipeline (\textit{i.e.}, the region proposal extraction step) by replacing the selective search module with another convolutional network called the Region Proposal Network (RPN), which shares the features with the detection network. RPN takes an input image and returns a set of rectangular object proposals, each with an object score.  Figure \ref{background}(c) is a flowchart of the RPN where k is the number of anchors. When 300 proposal regions per image are used, the processing frame rate reaches 5 frames per second (including all steps). Additionally, the convolution layers are shared between detection and region proposal networks. The R-FCN \cite{dai2016r} approach was then developed to circumvent the process of repeatedly applying per-region subnetworks by sharing almost all computations across the entire image, using fully convolutional networks. Fig. \ref{background} (d) shows the block diagram for this technique. Finally, FPN was used in \cite{lin2017feature} to improve the object detection performance especially for small size objects since it concatenates the information of the deeper and early layers together to produce a decision. A typical FPN is shown in Fig. \ref{background} (e). \\
\textbf{Single Stage Detectors}: YOLO \cite{redmon2016you} was the first proposed single-stage detector, which viewed the problem of object detection as a regression problem \textit{i.e.}, regressing the bounding box coordinates. Since the whole detection framework is performed in a single stage, the training process can be performed in an end-to-end manner. YOLO's first version achieved 45fps, making it suitable for real-time detection. However, the performance was relatively worse than its two-stage counterpart. As shown in Fig. \ref{background} (f), the algorithm divides the image into $S\times S$ grids and checks whether the center of each object lies within a grid cell. After that, the matched grid cell will regress the bounding box of the selected object in the grid. Finally, the overlapping bounding boxes are merged to produce the most plausible bounding boxes. The initial version of YOLO had strong spatial constraints, which made nearby objects difficult to detect. In order to address this problem and scale up the detection framework to a variety of objects, YOLOv2 was proposed in \cite{redmon2017yolo9000}. YOLO's localization error and low recall were identified by \cite{redmon2017yolo9000} as its most important limitations, which were addressed through batch normalization, high resolution classifiers, the use of anchor boxes instead of fully connected layers, and the use of clustering to determine the bounding box sizes as priors. A multi-scale prediction was used in YOLOv3 \cite{redmon2018yolov3},  to estimate bounding boxes at three different scales. A new network, called Darknet-53, has been proposed in \cite{redmon2018yolov3},  Which combines Darknet-19 and a residual network with 53 convolutional layers. In addition, the activation function of softmax has been replaced by logistic classifiers. YOLOv4 \cite{bochkovskiy2020yolov4} was built upon CSPDarknet53 on top of YOLOv3 and used Weighted-Residual-Connections (WRC), Cross-Stage-Partial-connections (CSP), Cross mini-Batch Normalization (CmBN), Self-adversarial-training (SAT) and  Mish-activation to improve the performance. The YOLO framework has been used to develop several other models, including \cite{jocher2020yolov5,long2020pp,wu2021yolop,ge2021yolox,chen2021you,wang2021you}. SSD \cite{liu2016ssd} is another single-stage detector that at first, was as accurate as the two-stage detectors while being much faster than its two-stage competitors. The core idea behind SSD is to determine the category scores and box offsets for a set of predefined bounding boxes using small convolutional filters on top of the feature maps. As shown in Fig. \ref{background}(g), various scales of feature maps have been used to perform the prediction. RetinaNet \cite{lin2017focal} was then proposed to alleviate the problem of class imbalance. In RetinaNet, a new focal loss focusing on hard examples was proposed by adding a multiplicative factor to the cross-entropy loss. Through this approach, the performance finally reached the performance of the SOTA two-stage methods. The structure of RetinaNet as shown in Fig. \ref{background}(h) uses the FPN as the neck of the pipeline.\\
The typical backbones used to extract learned features from image include: VGGNet \cite{simonyan2014very}, ResNet \cite{he2016deep},  ResNeXt \cite{xie2017aggregated}, Inception \cite{szegedy2016rethinking}, ZF Net \cite{zeiler2014visualizing} MobileNet \cite{howard2017mobilenets,howard2019searching}, DenseNet \cite{huang2017densely}, SqueezeNet \cite{iandola2016squeezenet}, ShuffleNet \cite{zhang2018shufflenet}, Darknet \cite{li2018detnet}, EfficientNet \cite{tan2019efficientnet} and Hourglass \cite{newell2016stacked}.
\section{Generic Small Object Detection}\label{generic}
Throughout this section, we will examine extensively SOD methods for both image and video modalities for generic applications. In Fig. \ref{tax}, we have categorized the methods for each modality and discussed how they are related below. 
 
\begin{figure*}[!h]
 {
 \includegraphics[width=0.9\textwidth]{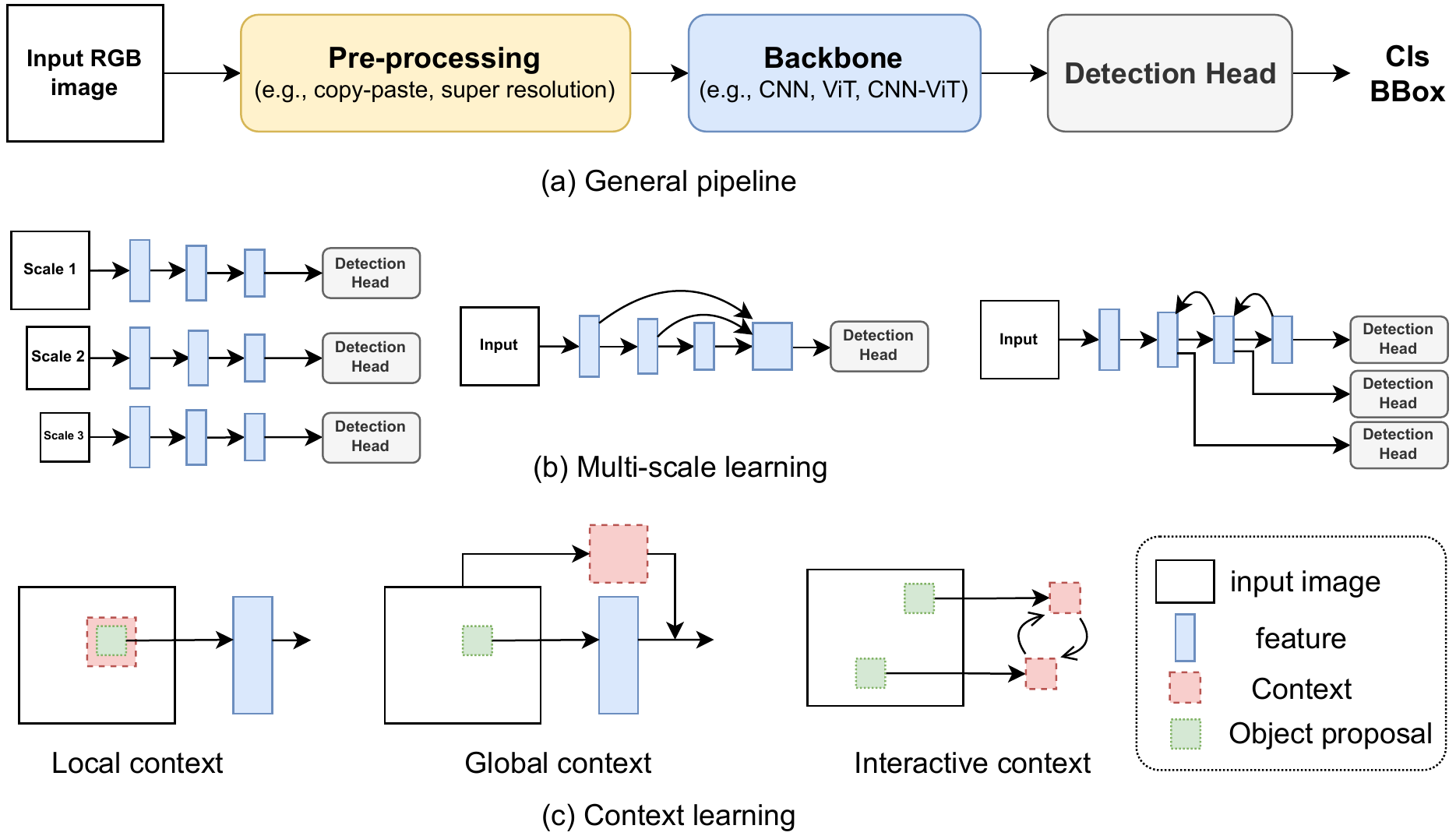}}
 \centering
 \caption{Block diagram of image-based SOD methods (for both maritime and generic applications).}
   \label{imagebased}
 \end{figure*}
\subsection{Image based SOD}
The topics covered in this section include training datasets, architecture, feature learning and objective loss functions. Fig. \ref{imagebased} shows a general block diagram of image-based SOD methods.
\subsubsection{Data Preparation}
\par\noindent\textbf{Data Augmentation.}
In computer vision, data augmentation is commonly used to address the problem of limited labelled data samples. Its goal is to generate a large, high-quality, and diverse set of training datasets that will enable deep learning models to be more robust and generalizable. The traditional methods of data augmentation can be broadly categorized into: (\textbf{i}) geometric transformations-based, including rotation, scaling, flipping, cropping, padding, translation, affine transformation, \textit{etc.} (\textbf{ii}) Photometric transformations-based, \textit{i.e.,} changing the color components, which include brightness, contrast, hue, saturation, \textit{etc.} In addition to these pixel-level adjustments based data augmentation methods, there are several patch-level manipulation methods, such as random erase~\cite{zhong2020random}, CutOut~\cite{devries2017improved}, CutMix~\cite{yun2019cutmix} and grid mask~\cite{chen2020gridmask}. Recent advances in Generative Adversarial Networks (GANs) provide a new avenue for data augmentation~\cite{antoniou2017data} by synthesizing realistic training samples of different styles~\cite{CycleGAN2017} or even novel unseen classes~\cite{wang2018low}. Moreover, Cubuk~\textit{et al.}~\cite{cubuk2019autoaugment} proposed a reinforcement learning-based data augmentation method, ``AutoAugment", to automatically search for the optimal augmentation strategy to train a classification model. Various data augmentation techniques have been used with the existing object detection methods, such as the horizontal flipping used with Fast R-CNN~\cite{girshick2015fast} and Cascade R-CNN~\cite{cai2019cascade}, saturation and exposure shifts used in YOLO~\cite{redmon2016you} and YOLO9000~\cite{redmon2017yolo9000}, and the ``Mosaic'' strategy proposed with YOLOv4~\cite{bochkovskiy2020yolov4}. Zoph~\textit{et al.}~\cite{zoph2020learning} extended AutoAugment~\cite{cubuk2019autoaugment} to the object detection task by performing the augmentation operations on the bounding boxes.
However, existing object detection methods generally perform worse on small objects, compared to medium or large objects. There are two main reasons: (\textbf{i}) there are much less images containing small objects in the training dataset, leading to a model that is biased towards medium or large objects; (\textbf{ii}) in those images containing small objects, the small object regions are too small, leading to a limited number of matched anchors. This namely decreases the probability of small objects to be detected. To address these problems, Kisantal~\textit{et al.}~\cite{kisantal2019augmentation} proposed two data augmentation methods accordingly. (\textbf{i}) An oversampling method was used to increase the number of training samples of small objects. (\textbf{ii}) To increase the number of small objects appearing in a single image, multi-copy-pasting of small objects was used to increase the likelihood of matching anchors with small target objects. Based on the copy-paste augmentation strategy~\cite{kisantal2019augmentation}, Chen~\textit{et al.}~\cite{chen2019rrnet} proposed an adaptive resampling augmentation method, which uses a pre-trained semantic segmentation model to determine suitable image regions for the augmented object pastes. This method effectively addresses the problems of background and scale mismatches when performing random pastes.  In order to exploit additional datasets of different object scale distributions to pre-train the network for small object detection, Yu~\textit{et al.}~\cite{yu2020scale} proposed a scale match approach to align the scale distributions of the pre-training dataset with that of the target small object dataset. Similarly to the Mosaic strategy~\cite{bochkovskiy2020yolov4}, Chen~\textit{et al.}~\cite{chen2020stitcher} proposed to balance the scale distribution of a training dataset by stitching multiple images of medium- or large-size objects to form a down-scaled collage image. Moreover, a feedback-driven decision paradigm based on the loss statistics of the minority small-scale objects was proposed to guide the image stitching process.

\par\noindent\textbf{Super Resolution.} The limited region of interest (RoI) for small objects results in insufficient feature information for an accurate detection prediction. To address this problem, a straightforward method is to perform super-resolution, namely recovering high-resolution images from their low-resolution counterparts~\cite{wang2020deep}. There are typically two types of super-resolution strategies for small object detection: (\textbf{i}) image super-resolution and (\textbf{ii}) feature super-resolution. Haris~\textit{et al.}~\cite{haris2021task} proposed to concatenate a super-resolution network prior to a detection network for an end-to-end training. The super-resolution process was also driven by the detection objectives, thus leading to better detection-oriented super-resolved images. Bai~\textit{et al.}~\cite{bai2018sod} proposed a multi-task generative adversarial network for small object detection (SOD-MTGAN). More specifically, SOD-MTGAN is composed of: (\textbf{i}) a generator which reconstructs super-resolved RoI images from the small blurred ones, and (\textbf{ii}) a multi-task discriminator to perform detection on the super-resolved RoI images and differentiates real high-resolution RoI images from the fake generated ones. Image super-resolution can help recover details of small objects in an image, thereby resulting in a moderate improvement in detection performance. However, image super-resolution based methods for small object detection suffer from several limitations. \textbf{Firstly}, super-resolving whole images can inevitably enlarge other irrelevant regions, which adversely impact detection performance. \textbf{Secondly}, if super-resolution is only performed on RoI images, object detection on the super-resolved RoI images will largely limit the detection performance due to the lack of context information. This second limitation can be alleviated by performing super-resolution on deep feature maps, which are generated by convolving context. Li~\textit{et al.}~\cite{li2017perceptual} proposed a Perceptual GAN to improve small object detection by generating the super-resolved features of small objects that cannot be discriminated from the features of large objects. Similarly, Noh~\textit{et al.}~\cite{noh2019better} used GAN to generate super-resolved features for small objects. This was shown to significantly improve the detection performance by providing a direct supervision to learning the super-resolved features of small objects using high-resolution features with appropriate receptive fields. In their article \cite{pang2019jcs}, Pang~\textit{et al.} introduced a unified network, called JCS-Net, to integrate the classification and super resolution tasks and to exploit the relationship between large and small scale objects (pedestrians) for recovering the detailed information.\\
Finally, several other methods perform semi-preprocessing steps to improve detection performance. For example, in \cite{ozge2019power} the authors used the overlapped tiling technique to increase the likelihood of small objects being present in the training stage. 

\subsubsection{Deep Learning Architecture}
\par\noindent\textbf{2D-CNN.} The majority of deep learning-based methods for detecting small objects rely on CNNs. 
These object detection methods can typically be categorized into anchor-based or anchor-free methods. 
Anchor-based methods primarily consists of two types of methods, namely, two-stage methods and one-stage methods (see Section~\ref{backgroundsec}). One-stage methods generally have a faster detection speed, while two-stage methods tend to have higher detection performance. 

Anchor-based two-stage object detection methods mainly consist of the following two stages: (\textbf{i}) a stage to generate object proposals from images; (\textbf{ii}) a stage to predict the final bounding boxes of objects from the region proposals. Representative two-stage CNN frameworks include: R-CNN~\cite{girshick2014rich}, SPPNet~\cite{he2015spatial}, Fast R-CNN~\cite{girshick2015fast}, Faster R-CNN~\cite{ren2015faster}, FPN~\cite{lin2017feature}, and Cascade R-CNN~\cite{cai2018cascade, cai2019cascade}. Anchor-based one-stage methods do not have a stage for generating region proposals. Instead, they directly generate the class probabilities of objects as well as the corresponding coordinates of the bounding boxes. Representative anchor-based one-stage methods include YOLO v1~\cite{redmon2016you}, SSD~\cite{liu2016ssd}, YOLO v2~\cite{redmon2017yolo9000}, RetinaNet~\cite{lin2017focal}, YOLO v3~\cite{redmon2018yolov3}, YOLO v4~\cite{bochkovskiy2020yolov4}, and YOLO v5~\cite{jocher2020yolov5} (see Section~\ref{backgroundsec}). 

Anchor-based methods usually have a large number of anchors and hyper-parameters, leading to a prohibitively high computation cost. To address these problems, recent anchor-free methods alleviate the need for anchors by performing detection through key-points. This largely reduces the number of hyper-parameters. Recent related works include CornerNet~\cite{law2018cornernet}, CenterNet~\cite{duan2019centernet}, FSAF~\cite{zhu2019feature}, FCOS~\cite{tian2019fcos}, and SAPD~\cite{zhu2020soft}.

\par\noindent\textbf{Image Transformer.} Several studies have suggested the use of transformers \cite{vaswani2017attention} for detecting objects following  Dosovitskiy~\textit{et. al.'s} pioneering work \cite{dosovitskiy2020image}. The Vision Transformer (ViT) was used for the first time in ViT-FRCNN \cite{beal2020toward} to examine the feasibility of transformers for complex object detection tasks. However, the SOD results revealed that the proposed method was not suitable and modifications were necessary to improve the detection performance. Moreover, the proposed method combines transformers and CNNs (\textit{i.e.}, does not merely use transformers). As a way to mitigate the reliance on CNNs and to propose a purely transformer-based object detection technique, You Only Look at One Sequence (YOLOS) was proposed in  \cite{fang2021you} to test the transferability of pre-trained transformers from image recognition to object detection. But YOLOS was unable to benefit from multi-scale features and achieved limited performance. With these limitations in mind, \cite{song2021vidt} proposed a method that integrates Vision and Detection Transformers (ViDT), and introduced three major contributions: \textbf{(i)} a new attention mechanism called Reconfigured Attention Module (RAM); \textbf{(ii)} a lightweight encoder-free neck architecture; and \textbf{(iii)} a token matching for knowledge distillation. 
\par\noindent\textbf{Mixed Architecture.}  The use of both CNNs and transformer architectures has been proposed in various studies. The Most common approach is to first use CNN networks as the backbone and extract several appropriate feature maps. Then these feature maps should be fed into a transformers for decision making. In the early work of transformer-based object detection (OD), Carion~\textit{et al.} \cite{carion2020end} proposed DEtection TRansformer (DETR) using transformers (with both encoder and decoder) on top of CNNs. DETR outperformed CNN-only based SOTA methods, while alleviating the need for complex post-processing steps such as Non-Maximum Suppression (NMS). Considering the computational cost of DETR, \cite{zhen2022deeply} proposed another compact end-to-end variant which represents the large weight matrix in one layer by low order matrices. Additionally, a decoder-only detector ($\text{D}^2$ETR) was proposed in \cite{lin2022d} to address complexity. Furthermore, two additional modifications of DETR were introduced in \cite{jiang2021guiding} in order to enhance learning and SOD performance.  \textbf{First}, in order to to update the positional information of the queries, a module called Guided Query Position (GQPos) was added to the decoder. \textbf{Second}, the authors proposed Similar Attention (SiA), a new fusion scheme that interpolates the low-resolution attention weight map to generate a high-resolution attention map, since multi-scale feature learning is computationally expensive. This idea was motivated from the fact that the relative positions of the objects is unique across different scales. A CNN-transformer based on deformable attention (following the idea of deformable convolution \cite{dai2017deformable}) and attending to just a small set  of sampling locations has been proposed by Zhu~\textit{et al.} \cite{zhu2020deformable}, which has the advantage of being trained much faster than DETR (with 10 times fewer training epochs). SOD performance was also improved by adding a multi-scale deformable attention module. Their method was referred to as ``Deformable DETR". Despite the fact that DETR and Deformable DETR only account for spatial information, they are still fast enough for Video SOD. A new method of extracting small-size features, SOF-DETR, has been proposed in \cite{dubey2021improving}, together with a normalized inductive bias. In a nutshell, SOF-DETR uses a multi-scale feature representation of the input image. Consequently, the input of the transformer captures richer information (both semantic and geometrical information) that is more suitable for SOD. Pre-training is performed only on the CNN block in DETR and Deformable DETR, but not on the transformer module. This was addressed by  \cite{dai2021up}, who proposed UP-DETR, which utilizes unsupervised pre-training for a pre-trained CNN backbone. However, since the pre-training of the transformer and CNN is done separately, they are unlikely to perform as well together. In FP-DETR \cite{wang2021fp}, the pre-training was thus performed on the encoder module (not the decoder) using ImageNet before fine-tuning the object detection task with a task adaptor. In \cite{wang2022resc}, a transformer-based object detection framework was proposed (RESC), which minimizes post-processing steps and the number of hyperparameters. RESC converges faster than DETR. In addition to being lighter, it enables the use of the FPN structure \cite{lin2017feature} to detect small objects.     

\subsubsection{Feature Learning}
\label{feature_learning}
\par\noindent\textbf{Multi-Scale Learning.} Multi-scale feature learning is one of the most common approaches for SOD, and several architectures have been developed to support it. Amudhan~\textit{et al.} \cite{amudhan2021rfsod} introduced RFSOD, a lightweight single-stage detector that can be used in embedded systems for real time applications. RFSOD's architecture is similar to that of the YOLO detector, and uses $3\times3$ and $1 \times 1$ convolutions for lightweight detection. By transferring and concatenating information from the earlier layers to the deeper layers, RFSOD increases the spatial resolution of the information in the last layers. This is critical for SOD and the concatenation is performed until that the receptive field reaches the size of $50 \times 50$, so that objects of size 32×32 and smaller can be detected. Chalavadi~ \textit{et al.} \cite{chalavadi2022msodanet} proposed mSODANET which consists of three main components: backbone network, Hierarchical Dilated Network (HDN), and Bi-directional Feature Aggregation Module (BFAM). EfficientNet \cite{tan2019efficientnet} was used to fully exploit the visual information contained in input images of varying sizes. Furthermore, the HDN was used to learn the contextual information of objects while the BFAM aims to resolve the network's limitation of top-down information flow (parallel connections from the last layers to the first layers) with cross-scale connections in order to improve the model efficacy. Fu~\textit{et al.} in \cite{fu2021small} extended the ResNet structure to ResNeXt-RC and proposed IIHNet. IIHNet is a convolution-based network based on three key concepts:  (\textbf{i}) information fusion; (\textbf{ii}) information exchange between different resolutions and modules; and (\textbf{iii}) a multi-scale network. Furthermore, \cite{he2021small} proposed a lightweight network known as YOLO-MXANet which uses a powerful backbone based on the MobileNext \cite{zhou2020rethinking}  named SA-MobileNeXt, as a mean to incorporate both spatial and channel attention. Along with the addition of another scale from the shallower layers to improve the performance of SOD, the number of parameters was markedly reduced from  61.5 M to 13.8 M. The authors in \cite{qi2022small} proposed a single stage SODNet composed of an adaptively spatial parallel convolution module (ASPConv) and a fast multi-scale fusion module (FMF) to optimize the spatial information extraction and to fuse the spatial and semantic information. By design, FMF preserves both spatial and semantic information. Following the SSD idea, Cui~\textit{et al.} \cite{cui2018mdssd} proposed  a Multi-scale  Deconvolutional  Single  Shot  Detector (MDSSD), where multiple feature maps at different scales are upsampled to increase the spatial resolution. For better localization of small objects, concatenation is used in \cite{liu2020small}, instead of summation in the fusion block to preserve more information across layers.

\par\noindent\textbf{Context Learning.} Objects are not isolated and they usually co-vary with other objects or particular backgrounds, which provides a rich source of contextual associations. 
For context learning, there are typically two types of approaches: (\textbf{i}) deep CNNs provide an \textit{implicit} way to model the spatial context for each pixel through the convolution and pooling operations. 
In order to incorporate the local context information, existing methods generally manually select the surrounding regions and aggregate their features to enhance the target regional feature~\cite{li2016attentive, chen2018context}.
In order to model the global context information, enlarging the receptive field to cover the whole image and performing global pooling is commonly employed. Besides, Bell~\textit{et al.}~\cite{bell2016inside} regarded feature maps as four sequences of feature maps arranged in the four cardinal directions, \textit{i.e.,} right, left, up and down, and proposed to model the global context information by using four recurrent neural networks (RNNs) to process each sequence and concatenating the outputs. 
To enhance the context learning of deep CNNs, a number of strategies have been developed to capture the multi-scale context~\cite{cui2020context,lim2021small} (See~\textbf{Multi-scale Learning} in Section~\ref{feature_learning}). Moreover, an attention mechanism has been used to effectively extract contextual information for object detection~\cite{li2016attentive, shen2019indoor}. (\textbf{ii}) Another line of methods involves \textit{explicitly} modeling the contextual information, such as scene-to-object and object-to-object relationships at the semantic level or in terms of the spatial layout. Fu~\textit{et al.}~\cite{fu2020intrinsic} proposed a context reasoning method for small object detection, which models the object-to-object relationships using the semantic features and the spatial geometric information (\textit{i.e.,} location, size, and aspect ratio) of object regions with a graph convolutional network (GCN). Using the learned contextual relations, the regional features were then updated for both classification and regression, resulting in improved performance for detecting small objects. Leng~\textit{et al.}~\cite{leng2021realize} proposed to model object-to-object relations and use the reliable object proposals with their pairwise relations to help classify and localize ambiguous object proposals.

\par\noindent\textbf{Region Proposal.} SOD performance of deep networks can be greatly enhanced by higher input image resolution. Using high-resolution data, however, requires considerably more computational power. To mitigate this bottleneck, one approach is to select the most promising regions and discard the rest of the input image. QueryDet was developed by Yang~\textit{et al.} \cite{yang2022querydet} which first localizes small objects roughly, then refers to high resolution feature maps for better adjustment of bounding box coordinates. Bosquet ~ \textit{et al.} \cite{bosquet2020stdnet} proposed STDnet which relies on two components: Region Context Network (RCN) and Region Of Interest (ROI) Collection Layer (RCL). As a result of processing only specific areas, high-resolution feature maps are kept in deeper layers, thereby increasing SOD performance. Additionally, in order to improve adaptation, both the number and the size of anchor boxes were learned by k-means in  \cite{bosquet2020stdnet}. In \cite{liu2021modified}, MdrlEcf was proposed as a way to exploit deep reinforcement learning (DRL) with a new reward function and an efficient attention network added to a CNN for the task of SOD with very high resolution remote sensing images. Based on FastMask \cite{hu2017fastmask}, Wilms~\textit{et al.} \cite{wilms2018attentionmask} proposed AttentionMask, a class-agnostic object proposal generation algorithm that is well suited for SOD. AttentionMask is biologically-inspired and includes scale-specific attention maps.     

\subsubsection{Loss Function Regularization}
While most existing methods focused on redesigning the neural network architecture or utilizing some prior information in order to boost SOD performance, fewer works employed different loss functions or added penalty terms to the classical loss functions in order to boost SOD performance. We can cite RetinaNet \cite{lin2017focal}, which is designed to focus on the most challenging samples (\textit{e.g.}, small objects) by multiplying a term proportional to the network's confidence into the classical cross-entropy loss. Other methods modify the standard IoU loss, including Intersection over Detection, Generalized IoU \cite{rezatofighi2019generalized}, Wasserstein distance \cite{wang2021normalized},
and Complete IoU \cite{zheng2021enhancing}; The detailed explanation of these methods can be found in Section 6.2.1.   
\begin{figure*}[]
	\centering
		 \includegraphics[width=0.8\linewidth]{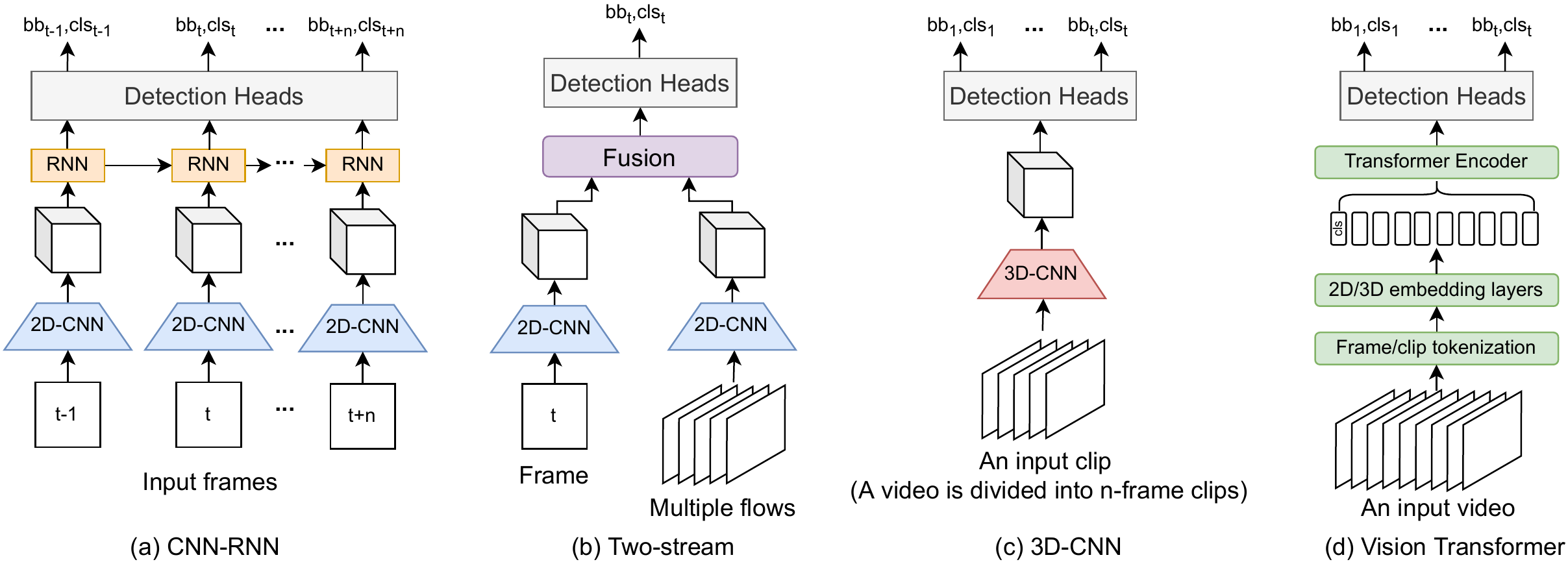}
	\caption{Typical commonly used structures for small object detection in videos.}
	\label{video-based}
\end{figure*}
\subsection{Video based SOD}
In general, videos provide additional temporal contextual information that is not contained in still images. Several previous methods exploit temporal information in an ad-hoc way~\cite{han2016seq,feichtenhofer2017detect}. These methods depend essentially on the static object detection results produced by an image-based object detector and then use the temporal information in a post-processing stage. This however leads to sub-optimal results since the training of the object detector does not take advantage of temporal information. More recent methods~\cite{xiao2018video,liu2018mobile} have incorporated the temporal information into training either by aggregating feature maps across different frames or by predicting object proposals between frames. As a result, the video object detection performance has been largely improved. With so many redundancies between adjacent frames, detection performance can be improved while still maximizing detection speed.
The use of temporal information can also improve detection performance when dealing with challenges such as motion blur, partial occlusion, small-scale objects, \textit{etc.} Our focus in this section is on the methods that jointly learn spatial and temporal information to detect small objects in video footage.
\subsubsection{Deep Learning Architecture}
In this section, we present general deep learning architectures (illustrated in Fig.~\ref{video-based}) for small object detection in videos.
\par\noindent\textbf{3D-CNN.} {While 3D-CNN is the easiest tool for integrating temporal and spatial information of video frames, it is rarely used for the task of object detection. In contrast, 3D-CNN has been deeply investigated for 3D object detection \cite{maturana2015voxnet}, action recognition \cite{ji20123d}, anomaly detection \cite{shin20203d}, \textit{etc.}  In contrast, of those limited studies, 3D-CNN was used in \cite{lin2019smoke} as a feature extractor in combination with Faster R-CNN in order to detect and localize smoke.     }

\par\noindent\textbf{RNN.} The recurrent neural network (RNN) is a type of neural network that processes temporal or time-series data. It has been widely used for video-based visual tasks following the pipeline shown in Fig.~\ref{video-based} (a). Tripathi~\textit{et al.}~\cite{tripathi2016context} proposed to use a recurrent neural network to extract the temporal context information, which is subsequently used to compute a regularization loss to better optimize the training of an object detector. Lu~\textit{et al.}~\cite{lu2017online} proposed Association LSTM, which is composed of an SSD and an LSTM networks. More specifically, SSD performs object detection on each frame. The features of the detected objects by SSD are stacked and then forwarded to the LSTM. An additional association error loss is applied to the LSTM outputs of two adjacent frames, to enforce the consistency of two neighboring frames in the temporal space. Compared to Association LSTM, which only uses limited motion information between two frames, Xiao~\textit{et al.}~\cite{xiao2018video} proposed a spatio-temporal memory network (STMN) to leverage the motion information across multiple frames. STMN is a bi-directional RNN, which is used to process the convolutional features of a sequence of multiple neighboring frames and also transfer the outputs to each frame. Therefore, the spatial and motion information of multiple neighbouring frames is all incorporated to compute the detection prediction for a target frame, thus effectively improving the detection performance. Moreover, to refine feature maps across frames, Liu~\textit{et al.}~\cite{liu2018mobile} proposed an inter-weaved recurrent-convolutional network, coined as Bottleneck-LSTM. By using depthwise separable convolutions and bottleneck design principles, Bottleneck-LSTM achieves a real time inference as well as a high detection performance.
\par\noindent\textbf{Video Transformer.} Due to their superior ability to detect long-range correlations, transformers have recently become very popular in object detection. Transformers have been applied to video based SOD to capture long term spatio-temporal dependencies. As described in \cite{he2021end} and \cite{zhou2022transvod}, TransVOD is the first end-to-end system for video object detection using spatio-temporal information. TransVOD uses multiple frames of the video as inputs to its spatial transformers, and uses another temporal transformer on top of it. These two transformers can link each object query and memory encoding outputs simultaneously. Two other extensions of TransVOD have been developed, called TransVOD++ and TransVOT Lite. TransVOD++ uses hard query mining (HQM) strategy to mitigate the redundancy of the number of objects and targets. Experiments show that the TransVOD framework can improve the performance of SOD. TransVOD++ is the first to achieve $90\%$ mAP on ImageNet VID dataset. The second extension, TransVOT was designed for real time object detection.    

\subsubsection{Spatio-Temporal Feature Aggregation}
In the previous section, we explained how sequence-based architectures such as 3D-CNN, RNN, and transformers have been applied to detect small objects. In other studies, the temporal and spatial features are mixed or aggregated during the process of object detection, \textit{e.g.}, by using 2D-CNN and finding the objects correlation over time. The STDnet-bST algorithm \cite{bosquet2020stdnet}  was proposed by Bosquet ~\textit{et al.} which first detects objects in frames using STDnet, and then links the detected objects using the Viterbi algorithm across the frames. In another extension, Bosquet~\textit{et al.} \cite{bosquet2021stdnet} proposed STDnet-ST, a spatio-temporal convolutional network method for SOD.  Built on STDnet, STDnet-ST operates on two consecutive frames simultaneously. These two frames are integrated together through a correlation module at shallower layers and a final tubelet linking module. The term ``tubelet linking" refers to forming sequences of the same objects across a video. Despite being based on the Viterbi algorithm, the tubelet linking module has three novelties, including \textbf{(i)} correlations are generated from the shallower layers of the convolution layers; \textbf{(ii)} to evaluate the degree of variability and confidence, a scoring system has been used; and \textbf{(iii)} dummy objects are introduced to suppress tubelets with incorrect data associations.       
A Faster R-CNN-like method called FANet was proposed by Cores~\textit{et al.} \cite{cores2020spatio} based on short-term spatio-temporal feature aggregation to produce first a detection set, followed by long-term object linking to refine the detection. They also introduced Tubelet Non-Maximum Suppression (T-NMS) to eliminated spatially redundant tubelets.    
\section{Maritime SOD}\label{maritime}
 
This section provides a literature review of SOD in maritime environments. Objects such as vessels, swimmers, obstacles, or plastic objects on the water's surface are included in this category.  


\subsection{Image based maritime SOD}
\label{maritime_sod}
This section is organized according to the flow of the detection pipeline shown in Fig. \ref{imagebased}. 
\subsubsection{Data Pre-processing} 

\textbf{Data augmentation.} Data augmentation is one of the most effective methods to improve the performance of small object detection. A number of data augmentation methods~\cite{kisantal2019augmentation} have been developed to increase the size and enrich the diversity of maritime training datasets, thus improving the robustness and the generalization ability of the detection models. 
In the maritime context, general data augmentation techniques, such as multi-angle rotation, color jittering, random translation, random cropping, horizontal flipping and adding random noises, have also been used in~\cite{you2019broad, liu2021enhanced, zhang2020intelligent, wang2021sdgh} to increase the diversity of samples.
In order to address the scarcity of real-world samples of small ships for training a deep learning based object detector, Chen~\textit{et al.}~\cite{chen2020deep} proposed to use a Gaussian Mixture Wasserstein GAN with Gradient Penalty (WGAN-GP) to generate synthetic small ships. Both real and synthetic data were used for training, significantly improving the detection performance over the case of not using synthetic data.  Moreover, Shin~\textit{et al.}~\cite{shin2020data} proposed a ``cut and paste" strategy to augment training images for maritime object detection. More specifically, the pre-trained mask-RCNN was used to extract the ship segments, which were then pasted in various background sea scenes to synthesize new images. The improved detection results confirmed the effectiveness of the synthetic ship images. Similarly, Hu~\textit{et al.}~\cite{hu2022somc} proposed a mixed strategy to mix the regions of sea surface objects with a number of varying scenes to increase the diversity and the number of training samples.\\
\textbf{Image Enhancement.} The complex marine environment makes maritime object detection challenging. The ocean wind,  waves, and currents usually cause marine object motion blur, which significantly degrades the performance of visual object detectors. Feng~\textit{et al.}~\cite{feng2021sharpgan} proposed ShapeGAN, a deblurring method based on GAN, which aims to remove motion blur from real sea images. The ship detection results of the sharp images are clearly superior to those of the blurred ones. In~\cite{tian2021image}, a GAN based low-quality to DSLR-quality image translator~\cite{ignatov2017dslr} was used to enhance the remote sensing ship imagery, leading to images with improved contrast and clarity. In~\cite{tian2021image}, the proposed image enhancement method was shown to improve detection performance, especially when training data is scarce. For image enhancement, deep learning is often combined with physical models. For instance, to improve maritime vessel detection, Guo~\textit{et al.}~\cite{guo2021lightweight} proposed a low-light image enhancement method based on deep learning and the Retinex theory~\cite{land1977retinex}. According to the Retinex theory, the observed image can be decomposed into reflectance and illumination components, so image quality can be improved by enhancing the illumination. 
To this end, Guo~\textit{et al.}~\cite{guo2021lightweight} proposed to learn a mapping between low-light images and their illumination-enhanced counterparts through a CNN-based model. This model was supervised by pairs of synthetic low-light and normal-light images. With the trained model, low-visibility maritime imagery was significantly enhanced, which improved the vessel detection in low-visibility environments. Similar maritime image enhancement methods have been proposed in~\cite{lu2021towards, yang2021deep}. The Atmospheric Scattering model~\cite{narasimhan2000chromatic} has also been used with deep learning to de-haze the maritime images to achieve an improved vessel detection performance in~\cite{guo2021heterogeneous}.\\ 
\textbf{Sea-Land Segmentation}. Another widely used pre-processing technique is sea-land segmentation or land masking. Usually, this technique is used when analyzing satellite images. Direct application of standard DNN-based methods in coastal areas, where the land and sea meet, can generate a high number of false positives due to similarities between urban structures and vessels. In order to reduce the false alarm rate, researchers used a pre-processing step in order to remove the land regions and thus reduce the amount of information for further analysis. Examples of DNN-based techniques include SeNet \cite{cheng2016senet}, which combines segmentation and edge detection methods in an end-to-end framework. Li~\textit{et al.} \cite{li2018deepunet}, developed DeepUNet, a pixel-level sea-land segmentation method based on U-Net. DeepUNet consists of a contracting path and an expansive path used to generate a high resolution optical output. Liu~\textit{et al.} \cite{liu2021laenet} proposed a lightweight multitask, end-to-end fully convolutional neural network without any down sampling to simultaneously segment the input image and extract edges from remote sensing images. In addition, a novel method (BS- Net) based on the joint learning network of boundary and segmentation is described in \cite{jing2021bs}, in which these two modules interact and enhance the sea-land segmentation result. In the literature, there are several other methods for separating sea from land, however since their details are beyond the scope of this survey, we do not elaborate further.

\subsubsection{Feature Learning}
\textbf{Multi-scale Learning.} 
Smaller objects have fewer pixels to work with compared to normal-size objects. Therefore, obtaining good representations of small objects can be challenging. Furthermore, after passing through a number of sub-sampling and striding operations, the top-layer feature maps may not include any features of small objects~\cite{liu2016ssd}. This makes detecting small objects more difficult. A multi-scale learning strategy is an effective method for improving the detection of small objects. It is also the most commonly used strategy for detecting maritime small objects.
\\
Multi-scale learning typically falls into two categories:
(\textbf{i}) multi-level features, \textit{i.e.,} combining features from different layers. Zhang~\textit{et al.}~\cite{zhang2019real} improved Faster R-CNN by fusing low- and high-level features to generate object proposals, predict bounding boxes and classification scores for float detection. Li~\textit{et al.}~\cite{li2021water} integrated feature maps from a number of layers by employing a feature pyramid network structure with deconvolutions into SSD, effectively improving the detection performance of remote objects in water surface. Additionally, the fusion of shallow features and deep features has also been used to detect ships in remote sensing images~\cite{zhang2020intelligent} for ship detection of remote sensing images. 
(\textbf{ii}) parallel multi-scale features, which are usually obtained by applying multiple parallel convolutions with different kernel sizes or dilated rates on the same input feature. 
Li~\textit{et al.}~\cite{li2018hsf} improved faster R-CNN by proposing a Hierarchical Selective Filtering (HSF) layer, which is composed of three parallel convolutional layers with kernel sizes $1\times1$, $3\times3$, $5\times5$, respectively. The HSF layer, which exploits features of multiple receptive fields, was used for both object proposal generation and bounding box regression, effectively detecting both inshore and offshore ships of varying sizes. Compared to the standard convolution, dilated convolution is more efficient since it enlarges the receptive field without increasing the number of parameters. 
Chen~\textit{et al.}~\cite{chen2021ship} proposed to enhance the feature representation of YOLOv3 by using multiple dilated convolutions to capture multi-scale context information for ship detection. Tian~\textit{et al.}~\cite{tian2021image} embeded multiple Atrous Spatial Pyramid Pooling (ASPP) modules in FPN to improve the detection performance for ships at different scales. Zhou~\textit{et al.} \cite{zhou2021image} proposed CRB-Net, a multi-scale image feature learning based method that can carry out adaptive weight adjustment (improved BIFPN) during feature fusion by attention mechanism and Mish activation (a novel self-regularized non-monotonic activation function \cite{misra2019mish}). Two SPPNets were also used to increase the receptive field of the features in layers 4 and 5 to isolate the most significant contextual features. The performance of CRB-Net was compared to 16 different deep learning-based methods for the detection of small objects on water surface, with promising results. 
\\

\par\noindent\textbf{Attention based learning.}
Multi-scale feature learning poses a challenge to real time object detection due to its increased complexity. This is because all areas in the input data (image/video) are exploited to localize objects. An alternative to reduce time and computational load is to use attention (whether spatially, temporally, or channel-wise) to eliminate irrelevant information and focus on that which is relevant to the object of interest.\\
For small object detection in maritime environments, Chen~\textit{et al.}~\cite{chen2021improved} proposed a single stage method, called ImYOLOv3 which integrates both spatial and channel attention modules (DAM) into a YOLOv3 network in order to better distinguish between ships and backgrounds. Their proposed end-to-end framework was successfully applied to optical remote sensing images. By adjusting receptive fields on three network branches, ImYOLOv3 achieved promising results for large, medium, and small sized objects. 
Nie~\textit{et al.}~\cite{nie2020attention} used both the channel attention modules and the spatial attention modules in a Mask-RCNN model to enhance the information propagation from the lower layers to the top layers. The use of the attention mechanism was shown to significantly improve the detection accuracy of small ship detection.  
Liu~\textit{et al.}~\cite{liu2021attention} used the Convolutional Block Attention Module (CBAM)~\cite{woo2018cbam}, which sequentially applies channel and spatial attention modules, to refine intermediate features of the object detection network. A similar attention mechanism was also used in~\cite{hu2021pag, fu2021improved, dong2021ship, li2021enhanced}.
Wang~\textit{et al.}~\cite{wang2021ship} used the Squeeze-and-Excitation (SE) attention module~\cite{hu2018squeeze} to dynamically perform channel-wise feature re-calibration, leading to an enhanced representational capacity of their detection network and an improved overall detection performance. A similar attention mechanism was also used in~\cite{hu2021ship}. Chen~\textit{et al.}~\cite{cheng2021robust} proposed a global attention module to adaptively fuse multi-modal features extracted from image and radar data for small floating waste detection~\cite{cheng2021flow}.


\subsubsection{Leveraging Segmentation methods}
\textbf{Foreground/Background Segmentation.} Saliency detection aims to mimic the low-level human visual attention mechanism, which localizes the most ``interesting" (salient) regions in an image for more efficient subsequent processing. Saliency object detection has been widely used in both traditional~\cite{sobral2015double, cane2016saliency} and deep learning-based~\cite{shao2019saliency} methods for maritime small object detection, to determine reliable object regions. More specifically, in~\cite{shao2019saliency}, saliency detection was applied on the predicted object proposal to refine their predicted locations for a more accurate ship detection.

\par\noindent\textbf{Semantic Segmentation.}  Smart modifications of the loss functions can result in a better feature representation for maritime small object detection. It was demonstrated in  \cite{moosbauer2019benchmark} that multitask (joint) learning, such as segmentation and object detection, can improve the performance of each task. A possible explanation is that due to joint learning, feature representation is no longer task-specific nor over-fitted to the training dataset. Cane~\textit{et al.}~\cite{cane2018evaluating} proposed the use of state-of-the-art deep semantic segmentation networks such as ENet \cite{paszke2016enet}, ESPNet \cite{mehta2018espnet} and SegNet \cite{badrinarayanan2017segnet} for maritime object detection. As a result of this, the segmentation stream improved greatly while the network needed fewer annotated, labelled images to train.       
Park~\textit{et al.}~\cite{park2022lightweight} proposed a lightweight Mask-RCNN by using an efficient backbone, \textit{i.e.,} MobileNetV2, to jointly perform warship detection and segmentation. 
To reduce the cost of dense pixel-level annotation, Zust~\textit{et al.}~\cite{vzust2022learning} proposed a weakly supervised method to train a semantic segmentation network for maritime obstacle detection.
\subsubsection{Generic OD for Maritime SOD}
Even though SOD in maritime environments presents some unique challenges in terms of shape and domain, several works have directly applied and evaluated generic object detection methods for this more challenging task. The main focus of these studies was to introduce a new maritime dataset and use the generic OD approaches as a baseline. This section reviews such prior works. YOLOv2 was evaluated by Lee~\textit{et al.} \cite{lee2018image} in maritime video surveillance with no changes to the overall network except a slight modification to the final layer used to classify objects into the 10 different ship classes. A speed of 30fps was achieved, thus making the method suitable for real time maritime detection. In \cite{moon2020comparative}, a cascade R-CNN \cite{cai2018cascade} with a HRNetV2 backbone for high resolution representation  \cite{wang2020deep2} was used to more accurately detect small objects in maritime environment. This accuracy was the consequence of maintaining information throughout all the layers. In another study, Shao~\textit{et al.} \cite{shao2018seaships} compared and analyzed the performance of Faster R-CNN (ZF Net, VGG16 Net, ResNet18, ResNet50, ResNet101), YOLO (DarkNet19), SSD (MobileNet, VGG16 Net) on their own maritime dataset. It was observed that YOLOv2 can achieve a proper trade-off between accuracy and speed in practical applications (average precision of 79 and speed of 91fps). The speed of YOLOv2 was adequate for real time video-based object detection. Aside from providing a new dataset for the maritime environment, the authors of  \cite{ribeiro2017data}, also used four different techniques (2 supervised and 2 unsupervised) to provide a benchmark for SOD in maritime environments. Parasad~\textit{et al.} \cite{prasad2018object} evaluated the performance of 23 classical and state-of-the-art Background Subtraction (BS) algorithms on visible range and near infrared range videos using the Singapore Maritime dataset. They found that those methods were not suitable for maritime environments (poor prediction), largely due to spurious dynamics of water, wakes, ghost effects, and multiple small detections for a single object. Therefore, BS methods must be adapted to suit the highly dynamic maritime backgrounds. The authors in \cite{scholler2019assessing} used LWIR input images, together with CNN-based methods such as RetinaNet (ResNet50), YOLOv3 (Darknet53) and Faster RCNN to localize objects at sea. In \cite{soloviev2020comparing}, the authors reported the results for Faster R-CNN, R-FCN and SSD on their own dataset. Compared to their other evaluated methods, Faster R-CNN with ResNet101 achieved the highest detection accuracy for large objects. Its accuracy was reduced, however, when they considered small objects. A cascading approach was used in  \cite{van2020automated} to monitor plastic pollution, using one network for the segmentation of regions of interest and another network for classification. In their comparison step, their goal was not to determine the exact location of the plastic bottles, but to predict their number in river streams. In \cite{chen2019port}, the YOLOv3 framework was used to accurately identify small, medium and large ships using three feature scales provided by DarkNet53. Varga~\textit{ et al.} \cite{varga2022seadronessee} presented a new sea-based vision dataset for identifying and localizing swimmers in open waters for emergency rescue missions. They compared the state-of-the-art CNN based techniques such as Faster R-CNN, CenterNet \cite{zhou2019objects}, and EfficientDet \cite{tan2020efficientdet} with different backbones and showed that Faster R-CNN with a deep network (ResNeXt-101-FPN) outperforms others.  However, it revealed very challenging to localize swimmers from a far distance, since they appear as points on the image.    
\subsubsection{Other Maritime SOD}
In \cite{li2018multiscale} the authors used a slightly different regression task by adding an angle parameter to the existing standard four bounding box parameters regression. This modification provides a more precise localization of rotated ships within a rectangular bounding box that is aligned with the ship’s direction. Similar approaches have been reported in \cite{qin2021mrdet,yi2021oriented,zand2021oriented}.  Using SSD, \cite{ghahremani2018cascaded} developed a cascade object detection method to identify obscure regions. Following some verification steps, the method considers the original high resolution input image (the one before down sampling) for decision making. This method does not require any modifications when different architectures are used. However, this cascading approach makes the method inappropriate for real time applications due to its high complexity.

\begin{figure*}[!h]
 \scalebox{1}{\begin{tabular}{c}
		  \includegraphics[scale=.82]{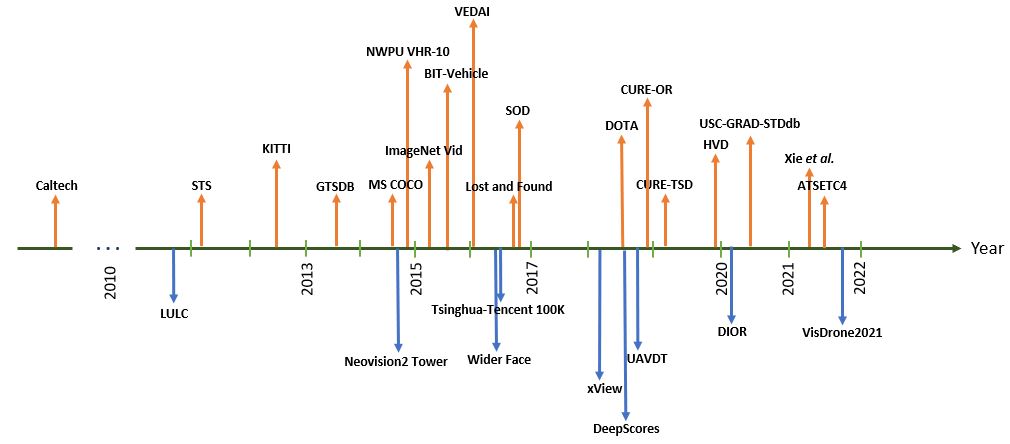}\\
		 (a)\\
		 \includegraphics[scale=.67]{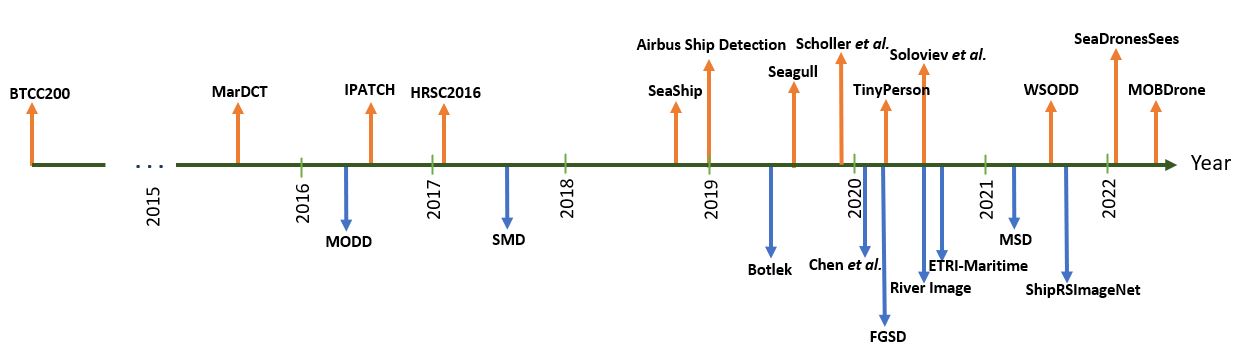}\\
		 (b)\\
	\end{tabular}}	
 \centering
 \caption{A brief chronology of SOD datasets, (a) for generic, and (b) for maritime environments. }
   \label{T1}
 \end{figure*}

\subsection{Video based maritime SOD}
Prior works for video-based maritime small object detection are typically categorized into: (\textbf{i}) spatial-based (\textit{i.e.,} frame-based) detection and (\textbf{ii}) spatio-temporal based detection. The first category of methods, \textit{e.g.,}~\cite{guo2021lightweight, guo2021heterogeneous, yang2021deep, lu2021towards, shao2019saliency,liu2021attention}, generally developed similar strategies compared to their image-based counterparts (see Section~\ref{maritime_sod}), and detected maritime small objects in videos frame by frame. While these methods (by only using the spatial information) have been able to achieve good detection accuracy and speed in several video based maritime applications, we believe that using the temporal information across video frames could lead to better performance by inferring relationships between moving objects. Therefore, this section focuses on the methods which leverage both the spatial and temporal information for maritime small object detection in videos.

Recent deep learning-based object detection methods generally perform well on large- and medium-sized objects. However, they perform poorly on small-sizes objects. Even though a number of specific techniques have been proposed to enhance the spatial features of small objects, their performance largely degrades in a dynamic environment characterized by background elements (\textit{e.g.}, water surface perturbations, sunlight reflection, floating driftwood and kelp), which are similar to the target objects in appearance or size. In such cases, the temporal information, \textit{i.e.,} the movement conveyed by multiple images/frames of the same scene, could be a useful cue to detect the presence of small objects. 
There are a number of works, which exploit both the spatial and temporal information for maritime small object detection in videos. Using the intersection of union of the bounding boxes between consecutive frames, Kim~\textit{et al.}~\cite{kim2018probabilistic} proposed to detect ships that could not be detected based solely on the spatial information of individual frames.
Marques~\textit{et al.}~\cite{marques2021size} proposed a Detector of Small Marine Vessels (DSMV), which exploited the temporal information to model backgrounds using a bi-directional gaussian mixture model. With the combination of DSMV and temporal information, the performance of general deep object detection methods was found to be significantly improved. The results confirm the effectiveness of using the temporal information for detecting maritime small objects in videos. Using a convolutional LSTM, Cruz~\textit{et al.}~\cite{cruz2019learning} extracted temporal features, which were combined with spatial features from CNNs, to detect objects in maritime airborne videos. Chen~\textit{et al.} \cite{chen2020video} proposed an automated ship recognition method consisting of four main steps: (\textbf{i}) feature extraction at different scales and construction of feature pyramids using ensemble YOLOv3 framework, (\textbf{ii}) bounding box generation, (\textbf{iii}) removal of interference bounding boxes using K-means algorithm and localization of ships, (\textbf{iv}) ship behavior analysis by a spatio-temporal constraints-based method on two consecutive frames. However, the reported spatio-temporal method still exhibits potential issues in handling fast moving ships and identifying individual ships in water-sky line as well as in dense (ship wise) environments such as ports and harbors. The components of the YOLOv3 network have been improved by Jie~\textit{et al.} \cite{jie2021ship} to achieve higher precision and recall values. Their contribution can be described as follows: (\textbf{i}) using the K-means algorithm to initialize the number of anchor boxes and their sizes based on the characteristics of the ships instead of the objects found in the VOC dataset, (\textbf{ii}) replacing the Sigmoid function with Softmax, (\textbf{iii}) introducing Soft Non-Maximum Suppression (Soft-NMS) to resolve the shortcomings of the standard NMS algorithm when  detecting overlapped objects. Finally, Deep Simple Online and Real time Tracking (Deep SORT) algorithm was used to accurately localize objects in frames with severe occlusions. They reported improvements of about 5\% and 2fps on average, in mean average precision (mAP) and in the number of analyzed Frame Per Second (FPS), respectively. An innovative spatio-temporal object detection method based on high-quality region proposals mainly centered around rigid (\textit{i.e.}, potential object) video locations is proposed in \cite{marie2018real}. The high quality regions of proposals were obtained by assessing textural variations at key video locations using a long-term keypoint tracking algorithm. Scale Invariant Feature Transform (SIFT) \cite{lowe2004distinctive} was shown to perform best compared to other keypoint extractors in terms of both accuracy and repeatability.    


\section{Evaluation of Small Object Detection}\label{dataset}
\subsection{Small Object datasets} 
A review of existing SOD datasets is provided in this section, along with an introduction to their characteristics. Datasets are categorized into two sets, generic datasets and maritime datasets. These datasets are summarized in Tables \ref{tab2} and \ref{tab3} and their chronological order is shown in Fig. \ref{T1}.  \\
\subsubsection{Generic SOD Datasets:}
\textbf{MS COCO} \cite{lin2014microsoft} The Microsoft Common Objects in COntext (MS COCO) dataset consists of images of complex everyday scenes that contain common objects in their natural settings. Despite not being specifically designed for SOD, MS COCO's average object size is smaller than most other well-known datasets, such as PASCAL VOC and ImageNet. Multiple SOD methods trained and tested their algorithms on a subset of MS COCO dataset that satisfies the definition of small objects  (\textit{i.e.}, less than $32 \times 32$ pixels).\\
\textbf{ImageNet Vid} \cite{russakovsky2015imagenet} is a large-scale dataset that was also not designed for SOD. Nevertheless, a number of detection frameworks reported their small object performance on a subset that consists of small objects. \\
\textbf{Lost and Found} \cite{pinggera2016lost} is the first publicly available lost-cargo dataset for the detection of small obstacles on the road. Thirteen challenging street scenarios were recorded, as well as 37 types of obstacles. The featured objects vary in size, color, material, and distance from the camera. Annotations  are  provided  for  every  10th frame of the videos.\\
\textbf{Swedish  Traffic  Signs  (STS)} dataset \cite{larsson2011using} was compiled by recording over 350 km of Swedish highways and city roads. The car was equipped with a camera with a focal length of 6.5 mm and a field of view of 41 degrees, which was pointing slightly to the right to capture road signs. Annotations  are  provided  for  every  5th  frame of the videos, which were recorded each time a sign appeared. Labeled  objects  include sign types, such as pedestrian crossing, designated lanes, no  standing  or  parking,  priority  road,  give  way, and signs that indicate a speed limit of 50  kph or 30 kph.\\
\textbf{Tsinghua-Tencent  100K} \cite{zhu2016traffic} With more than 100K images taken from 300 Chinese cities' road networks, this dataset is one of the most challenging datasets. A number of pre-processing techniques were applied to improve the quality of the images, including exposure adjustment.\\
\textbf{GTSDB} \cite{houben2013detection} The German Traffic Sign Detection Benchmark (GTSDB) is an image-based dataset with scenarios such as rural, urban, and highway driving, where most of the traffic signs occur only once. The images were selected from recordings near Bochum, Germany. \\
\textbf{CURE-TSD} \cite{temel2019challenging} is another sign detection dataset that provides a broad range of variations in illumination, occlusion, shadow, blur, or reflection. This dataset is relatively large and useful for the training of large deep learning models.\\
\textbf{Small Object Dataset (SOD)} \cite{chen2016r} is a subset of both MS COCO and Scene UNderstanding (SUN) datasets \cite{xiao2016sun}. The authors manually selected ten categories of objects which appear really small in the images.\\
\textbf{CURE-OR} \cite{temel2018cure} or Challenging  Unreal  and  Real  Environments  for  Object  Recognition(CURE-OR) contains objects with different  sizes, colors, and texture that are arranged in five different orientations. Images are acquired by five devices (iPhone 6s, HTC  One  X,  LG  Leon,  Logitech  C920  HD  Pro  Webcam, and  Nikon  D80) in both real-world (real)  and  studio  (unreal) environments. Despite the fact that this dataset was not specifically designed for SOD, it contains a large number of small objects, making it suitable for training and testing SOD methods.\\
\textbf{WIDER FACE} \cite{yang2016wider} is one large-scale face image dataset which contains 10 times more images than the other face detection datasets at the time of its release. Images were selected from the publicly available WIDER dataset \cite{xiong2015recognize}.\\
\begin{table*}
	\caption{Commonly used datasets for Generic SOD.}
	\label{tab2}
	\centering
	\scalebox{0.68}{\begin{tabular}{|c||c|c|c|c|c|c|c|c|p{2cm}|}
		\hline
		\textbf{Dataset} & \textbf{Application} & \textbf{Video}&\textbf{Image}  & \textbf{Shooting Angle (Type)} & \textbf{Resolution (pixels)}& \#\textbf{Object Classes} & \#\textbf{Instances} & \#\textbf{Image/Video}&\textbf{Public?}  \\
		\hline
		\hline	
			\textbf{MS COCO} \cite{lin2014microsoft} & Generic &   & \checkmark & (RGB) & NF &\makecell{91 Stuff C.\\80 Object C.} & 2.5M  &328K & Yes: \href{https://cocodataset.org}{\textcolor{green}{Click Here}}  \\
			\hline
			\textbf{ImageNet Vid} \cite{russakovsky2015imagenet} &Generic&\checkmark&&(RGB)&--&30&--&\makecell{4417\\($>$1.2M frames)}&Yes: \href{https://www.kaggle.com/competitions/imagenet-object-localization-challenge/data}{\textcolor{green}{Click Here}}\\
			\hline
		\textbf{Lost and Found} \cite{pinggera2016lost} & \makecell{Generic\\ (Autonomous Driving)} &  \checkmark &  & \makecell{On-board\\(stereo RGB sequence)} & $2048\times1024$ & 37 &--&\makecell{112\\ (2104 annotated frames)}&Yes: \href{https://www.6d-vision.com/6d-vision-powers-autonomous-driving}{\textcolor{green}{Click Here}} \\
		\hline
		\textbf{STS} \cite{larsson2011using}&\makecell{Generic\\ (Autonomous Driving)}&\checkmark&&\makecell{On-board\\(RGB)} &--&7&3488& \makecell{$>$20K frames\\ (20\% labeled)}   & Yes: \href{https://www.cvl.isy.liu.se/research/}{\textcolor{green}{Click Here}} \\
		\hline
		\textbf{Tsinghua-Tencent  100K} \cite{zhu2016traffic} &\makecell{Generic\\ (Autonomous Driving)}&&\checkmark&\makecell{On-board\\ Shoulder-mounted\\(panoramas RGB)}&$2048\times2048$&45&30K&100K&Yes: \href{https://cg.cs.tsinghua.edu.cn/traffic-sign/}{\textcolor{green}{Click Here}}\\
		\hline
		\textbf{GTSDB}\cite{houben2013detection} &\makecell{Generic\\ (Autonomous Driving)}&&\checkmark&\makecell{On-board\\ (RGB)}&$1360\times800$&4&1206&900&Yes: \href{https://benchmark.ini.rub.de/}{\textcolor{green}{Click Here}}\\
		\hline
		\textbf{CURE-TSD} \cite{temel2019challenging}&\makecell{Generic\\ (Autonomous Driving)}&\checkmark&&\makecell{On-board\\(RGB)}&$1628\times1236$&14&2.2M&\makecell{5733\\
		(1.7M frames)}&Yes: \href{https://github.com/olivesgatech/CURE-TSD}{\textcolor{green}{Click Here}}\\
		\hline
		
				\textbf{SOD} \cite{chen2016r} &Generic&&\checkmark&(RGB)&--&10&8393&4925&--\\
		\hline
		
			\textbf{CURE-OR} \cite{temel2018cure}&Generic&&\checkmark&(RGB)&NF&100&--&1M&Yes: \href{https://github.com/olivesgatech/CURE-OR}{\textcolor{green}{Click Here}}\\
		\hline
		
			\textbf{WIDER FACE} \cite{yang2016wider}&\makecell{Generic\\(Face Detection)}&&\checkmark&(RGB)&--&60&393K&32.2K&Yes: \href{http://shuoyang1213.me/WIDERFACE/}{\textcolor{green}{Click Here}} \\
		\hline
		
			\textbf{DeepScores } \cite{tuggener2018deepscores}&\makecell{Generic\\(optical Music Recognition)}&&\checkmark&(GS)&$1894 \times 2668$&123&80M&300K&Yes: \href{https://tuggeluk.github.io/deepscores/}{\textcolor{green}{Click Here}}\\
		\hline
		\textbf{ATSETC4} \cite{liang2021small}&\makecell{Generic\\(Air-Target Recognition)}&\checkmark&&(RGB)&--&4&--&\makecell{2400\\(60K frames)}& Yes \\
		\hline
		
		\textbf{HVD} \cite{song2019vision}&\makecell{Generic\\(Vehicle Detection)}&&\checkmark&(RGB)&$1920 \times 1080$&3&57290&11129&Yes: \href{https://drive.google.com/file/d/1li858elZvUgss8rC_yDsb5bDfiRyhdrX/view}{\textcolor{green}{Click Here}} \\
		\hline
		
		\textbf{BIT-Vehicle} \cite{dong2015vehicle}&\makecell{Generic\\(Vehicle Detection)}&&\checkmark&(RGB)&\makecell{$1600\times1200$\\$1920\times1080$}&6&&9850&Yes\\
		\hline
		
		\textbf{KITTI} \cite{geiger2012we}&\makecell{Generic\\(Autonomous Driving)}&&\checkmark&(RGB)&--&2&$>$100K&80256&Yes: \href{http://www.cvlibs.net/datasets/kitti/}{\textcolor{green}{Click Here}}\\
		\hline
		
		\textbf{Caltech} \cite{dollar2009pedestrian}&\makecell{Generic\\(Pedestrian Detection)}&\checkmark&&(RGB)&$640 \times 480$&3&350K&\makecell{1M frames\\(250K labeled frames)}&Yes: \href{http://www.vision.caltech.edu/Image_Datasets/CaltechPedestrians/}{\textcolor{green}{Click Here}}\\
		\hline
		
		\textbf{USC-GRAD-STDdb} \cite{bosquet2020stdnet}&\makecell{Generic}&\checkmark&&(RGB)&$1280\times720$&5&56K&\makecell{115\\($>$25K frames)}& Yes: \textcolor{green}{Under Request}\\
		\hline
		
		\textbf{UAVDT} \cite{du2018unmanned}&\makecell{Generic\\(Vehicle Detection)}&\checkmark&&\makecell{UAV based\\ (RGB)}&$1080\times540$&3&841.5K&\makecell{100\\(80K frames)} &Yes: \href{https://sites.google.com/view/grli-uavdt}{\textcolor{green}{Click Here}}\\
		\hline
		
		\textbf{VisDrone2021} \cite{zhu2020detection}&\makecell{Generic}&\checkmark&\checkmark&\makecell{UAV based\\ (RGB)}&\makecell{Image:$2000\times1500$\\Video:$3840\times2160$}&10&$>$2.6M&\makecell{400 Videos, $>$10K Imgages\\($>$265K frames)}&Yes: \href{http://www.aiskyeye.com/}{\textcolor{green}{Click Here}}\\
		\hline
		
		\textbf{Neovision2 Tower} \cite{khosla2014neuromorphic}&\makecell{Generic}&\checkmark&&\makecell{On-board\\(RGB)}&$1920\times1080$&5&--&\makecell{100}&Yes: \href{http://ilab.usc.edu/neo2/dataset/}{\textcolor{green}{Click Here}}\\
		\hline
		
		\textbf{NWPU VHR-10} \cite{cheng2014multi}&\makecell{Generic}&&\checkmark&\makecell{Satellite based\\(RGB\&CIR)}&--&10&--&800& Yes: \href{http://pan.baidu.com/s/1c0w8h3q}{\textcolor{green}{Click Here}}\\
		\hline
		
		\textbf{LULC} \cite{yang2011spatial,yang2010bag}&\makecell{Generic}&&\checkmark&\makecell{Satellite based\\(RGB)}&$256\times 256$&21&--&2100&Yes: \href{http://vision.ucmerced.edu/datasets/}{\textcolor{green}{Click Here}}\\
		\hline
		
	\textbf{DOTA} \cite{xia2018dota,ding2021object}&\makecell{Generic}&&\checkmark&\makecell{Aerial \& Satellite Images\\ (RGB)}&\makecell{From\\$800\times 800$ to \\ $20000 \times 20000$}&18&$>$1.7M&11268&Yes: \href{https://captain-whu.github.io/DOTA/}{\textcolor{green}{Click Here}}\\
		\hline
		
	\textbf{Xie~\textit{et al.}} \cite{xie2021small}&\makecell{Generic\\(Drone Detection)}&\checkmark&&(RGB)&\makecell{$1920\times1080$\\$2048\times 1538$ \\ $4096 \times 1800$}&2&--&6&No\\
		\hline
   \textbf{xView} \cite{lam2018xview}&Generic&&\checkmark&\makecell{Satellite based\\ (RGB)}&$1500\times1200$&60&1M&1413&Yes: \href{http://xviewdataset.org/}{\textcolor{green}{Click Here}}\\
   \hline
   \textbf{VEDAI} \cite{razakarivony2016vehicle}&\makecell{Generic\\(Vehicle Detection)}&&\checkmark&\makecell{Aerial based\\ (RGB \& NIR)}&$1024\times1024$&9&3640&1210&Yes \href{https://downloads.greyc.fr/vedai/}{\textcolor{green}{Click Here}}\\
   \hline
	\textbf{DIOR} \cite{li2020object}&Generic&&\checkmark&\makecell{Satellite based\\ (RGB)}&$800\times800$&20&192472&23463&Yes:\href{http://www.escience.cn/people/gongcheng/DIOR.html}{\textcolor{green}{Click Here}}\\
   \hline
	\end{tabular}}
\end{table*}
\textbf{DeepScores } \cite{tuggener2018deepscores} is an annotated dataset that contains high quality images of thousands of musical scores, partitioned into 3000000 sheets of written music with symbols of varying shapes and sizes. In addition to being unique, this dataset is the largest public dataset with close to a hundred million small objects (\textit{i.e.}, musical scores).\\
\textbf{ATSETC4} \cite{liang2021small} dataset contains small video clips selected from real-captured videos from the internet in various locations and conditions such as fields, cities, virtual environments and complex weather conditions. A total of four types of flying objects were included in this dataset: birds, fire balloons, fixed-wing UAVs, and rotor UAVs.\\
\textbf{Highway Vehicle Dataset (HVD)} \cite{song2019vision} includes images captured from the video monitoring of highway in Hangzhou, China. The images were captured by 23 surveillance cameras. There are three object classes: bus, car, and truck.\\
\textbf{BIT-Vehicle} \cite{dong2015vehicle} dataset contains images displaying changes in illumination conditions, vehicle scale, vehicle color and viewpoints. The following classification labels have been adopted: bus, microbus, minivan, sedan, SUV, and truck. There are 150 different vehicles in each category.\\
\textbf{KITTI} \cite{geiger2012we} is a well-known dataset for autonomous driving and vehicle detection. There are 7418 training images and 7518 testing images with 2D and 3D bounding boxes, along with a bird's eye view bounding box for evaluation. There are three categories of samples in the dataset: easy, moderate, and hard.\\
\textbf{Caltech Dataset} \cite{dollar2009pedestrian} is challenging because it includes objects that are frequently occluded and have low resolutions. The data was acquired by a vehicle travelling in regular traffic in an urban environment for approximately ten hours. The 30Hz videos were captured in the greater Los Angeles metropolitan area, which has a high pedestrian density.\\
\textbf{USC-GRAD-STDdb} \cite{bosquet2020stdnet} is a YouTube video dataset for small objects. It includes air,  land, and sea landscapes with the following objects: drone, bird, boat, vehicle, and person.\\
\textbf{UAVDT} \cite{du2018unmanned} is a large-scale UAV-based video dataset designed for vehicles detection and tracking. There are bounding boxes as well as useful information such as vehicle category, occlusion, and weather condition included in this manually annotated dataset. The videos were extracted from 10 hours raw videos. \\
\textbf{VisDrone2021} \cite{zhu2020detection} is a drone-based dataset collected by the AISKYEYE team at Tianjin University, China. This dataset covers 14 different cities in both urban and country areas. Object types in the dataset include pedestrians, vehicles, bicycles, \textit{etc.}\\
\textbf{Neovision2 Tower} \cite{khosla2014neuromorphic} includes videos captured from a fixed camera mounted atop Stanford University's Hoover Tower. This project was funded by the Defense Advanced Research Projects Agency (DARPA) under the Neovision2 program.\\
\textbf{NWPU VHR-10} \cite{cheng2014multi} is a high spatial resolution remote sensing
image dataset containing 10 classes of objects (airplanes, ships, storage
tanks, baseball diamonds, tennis courts, basketball courts, ground track
fields, harbors, bridges, and vehicles). Images were acquired from the Google Earth and Vaihingen datasets \cite{cramer2015dgpf}.\\
\textbf{LULC} \cite{yang2011spatial,yang2010bag} or land use/land cover is a publicly available remotely sensed dataset with 21 classes of agricultural land, airplanes, baseball diamonds,
beaches, buildings, chaparrals, dense residential areas, forests, freeways, golf
courses, harbors, intersections, medium density residential areas, mobile
home parks, overpasses, parking lots, rivers, runways, sparse residential areas,
storage tanks, and tennis courts. \\
\textbf{DOTA} \cite{xia2018dota,ding2021object} or Dataset for Object deTection in Aerial Images is a large-scale dataset containing objects of different scales, orientations and shapes. Objects include planes, ships, storage tanks, baseball diamonds, tennis courts, swimming pools, ground track fields, harbors, bridges, large vehicles, small vehicles, helicopters, roundabouts, soccer ball fields, container cranes, airports, helipads and basketball courts. \\
\textbf{Xie~\textit{et al.} Dataset} \cite{xie2021small} is a video dataset acquired with 3 different EO cameras.  \\
\textbf{xView Dataset} \cite{lam2018xview} is a dataset collected from WorldView-3 satellite with a spatial resolution of 0.3 m. \\
\textbf{VEDAI Dataset} \cite{razakarivony2016vehicle} is an aerial image dataset consisting of nine classes including boats, cars, camping cars, planes, pick‐ups, tractors, trucks, vans, \textit{etc.} The images were acquired by Utah AGRC with a resolution of 0.125 m. \\
\textbf{DIOR} \cite{li2020object} is a large-scale public dataset for object detection in optical remote sensing images. It includes a wide range of objects with inter- and intra-class variabilities.  \\
\subsubsection{Maritime Datasets:}
\textbf{TinyPerson} \cite{yu2020scale} image dataset is a collection of selected images taken from maritime videos uploaded on the internet.   \\
\textbf{Scholler~\textit{et al.} Dataset} \cite{scholler2019assessing} contains more than $20K$ Long Wavelength Infrared images acquired from ferries in the near coastal area of southern Funen archipelago. The images were acquired with a camera facing the direction of travel. Boats and buoys are the two main classes in the annotation.\\
\textbf{HRSC2016 Dataset} \cite{liu2017high} or High Resolution Ship Collection is one of the earliest publicly available datasets for ship recognition. It includes Google Earth images with standard bounding boxes, ship head positions, and rotated bounding boxes with information about ship types and categories. Image resolutions range from 0.4m to 2m.\\
\textbf{ETRI-Maritime Dataset} \cite{soloviev2020comparing} is a collection of RGB images captured, purchased, and collected from the Internet. There are 12 types of ships and buoys in the dataset, including buoys, fishing boats, cruise ships, ferries, container ships, gas carriers, other cargo ships, tugboats, barges, coast guards, warships, and yachts.\\
\textbf{SeaShip Dataset} \cite{shao2018seaships} is a large-scale dataset of images of six types of ships namely ore carriers, bulk cargo carriers, general cargo ships, container ships, fishing boats, and passenger ships. This dataset is not specifically designed for small objects. However, a large proportion of its objects are long-range, making it suitable for SOD. The images were selected from more than 10K video segments captured by surveillance system installed along the coastline of Hengqin Island, Zhuhai city, China.\\
\textbf{WSODD Dataset} \cite{zhou2021image} or Water Surface Object Detection dataset was developed for obstacle detection on water surfaces. The images include oceans, rivers, and lakes that were acquired at different times and weather conditions, such as during the day, twilight, and night, sunny, cloudy, or foggy conditions. Object classes include boats,
ships, balls, bridges, rocks, persons, rubbish, masts, buoys, platforms,
harbors, trees, grasses, and animals.\\
\textbf{Seagull Dataset} \cite{ribeiro2017data} is a dataset representing challenging maritime scenarios, similar to real world
scenarios. Glare, wave crests, wakes, and variations of perspective are all evident in this dataset. The recording was performed with an Alfa extended UAV built and designed by the Portuguese Air Force Research Center for research purposes. \\
\textbf{Soloviev~\textit{et al.} Dataset} \cite{soloviev2020comparing}
includes two different datasets: one with images from 135 videos captured from a watercraft moving between the cities of Turku and Ruissalo in South-West Finland, and the other has data continuously collected from two sensors in various geographic and environmental conditions.\\
\textbf{River Image Dataset} \cite{van2020automated} was collected by cameras installed on bridges at five water ways in Jakarta, Indonesia for monitoring plastic pollution.\\
\begin{table*}
	\caption{Commonly used datasets for Maritime SOD.}
	\label{tab3}
	\centering
	\scalebox{0.7}{\begin{tabular}{|c||c|c|c|c|c|c|c|c|p{2cm}|}
		\hline
		\textbf{Dataset} & \textbf{Application} & \textbf{Video}&\textbf{Image}  & \textbf{Shooting Angle (Type)} & \textbf{Resolution (pixels)}& \#\textbf{Object Classes} & \#\textbf{Instances} & \#\textbf{Image/Video}&\textbf{Public?}  \\
		\hline
		\hline	
	\textbf{TinyPerson} \cite{yu2020scale}&\makecell{Maritime\\(Person Detection)}&&\checkmark&\makecell{UAV based\\ (RGB)}&\makecell{From\\$497\times 700$ to \\ $4064 \times 6354$}&2&$>$72K&\makecell{2369\\(1610 labeled)}& Yes: \href{https://github.com/ucas-vg/TinyBenchmark}{\textcolor{green}{Click Here}}\\
		\hline
		
	\textbf{\textbf{Scholler~\textit{et al.}}} \cite{scholler2019assessing}&\makecell{Maritime\\(Ship Detection)}&&\checkmark&\makecell{On-board\\(LWIR)}&$640\times480$&2&--&$>$21k&No\\
		\hline
		
	\textbf{HRSC2016} \cite{liu2017high}&\makecell{Maritime\\(Ship Detection)}&&\checkmark&\makecell{Satellite based\\(RGB)}&\makecell{From\\$300\times 300$ to \\ $1500 \times 900$}&25&2976&1061&Yes:\href{https://www.kaggle.com/guofeng/hrsc2016?select=HRSC2016_dataset.zip}{\textcolor{green}{Click Here}}\\
		\hline
		
	\textbf{ETRI-Maritime} \cite{soloviev2020comparing}&\makecell{Maritime\\(Ship Detection)}&&\checkmark&\makecell{(RGB)}&NF&12&50K&37694&No\\
		\hline
		
	\textbf{SeaShip} \cite{shao2018seaships}&\makecell{Maritime\\(Ship Detection)}&&\checkmark&\makecell{Shore based\\(RGB)}&$1920\times1080$&6&40077&31455&--\\
		\hline
		
	\textbf{WSODD} \cite{zhou2021image}&\makecell{Maritime\\( Obstacle Detection)}&&\checkmark&\makecell{(RGB)}&$1920\times1080$&14&21911&7467&Yes: \href{https://github.com/
sunjiaen/WSODD; https://github.com/sunjiaen/BTRDA}{\textcolor{green}{Click Here}}\\
		\hline
		
	\textbf{Seagull} \cite{ribeiro2017data}&\makecell{Maritime\\(Ship Detection)}&\checkmark&&\makecell{UAV based\\(RGB\&NIR\&IR\&Hyperspectral)}&\makecell{$1920\times1080$\\$1024\times768$\\$640\times480$\\$384\times288$\\$1024\times648$}&6&--&\makecell{19\\(150K frames)}&Yes:\textcolor{green}{Under Request} \href{https://vislab.isr.tecnico.ulisboa.pt/seagull-dataset/}{\textcolor{green}{Click Here}}\\
		\hline
		
	\textbf{Soloviev~\textit{et al.}} \cite{soloviev2020comparing}&\makecell{Maritime\\(Ship Detection)}&&\checkmark&\makecell{Waterborne\\(RGB)}&$1920\times720$&--&850&400&No\\
		\hline
	\textbf{Soloviev~\textit{et al.}} \cite{soloviev2020comparing}&\makecell{Maritime\\(Ship Detection)}&&\checkmark&\makecell{Waterborne\\(RGB\&IR Thermal)}&$1200\times400$&4&9137&1750&No\\
		\hline
		
	\textbf{River Image} \cite{van2020automated}&\makecell{Maritime\\(Plastic Monitoring)}&&\checkmark&\makecell{(RGB)}&--&2&14968&1272&--\\
		\hline
		
	\textbf{SMD} \cite{prasad2017video}&\makecell{Maritime\\(Ship Detection)}&\checkmark&&\makecell{Shore based (RGB)\\On-board (RGB)\\Shore Based (NIR)}&$1920\times 1080$&10&240842&\makecell{81\\(31653 frames)}&Yes: \href{https://sites.google.com/site/dilipprasad/home/singapore-maritime-dataset}{\textcolor{green}{Click Here}}\\
		\hline
		
	\textbf{MarDCT} \cite{Bl-Io-Pe-15}&\makecell{Maritime\\(Ship Detection)}&\checkmark&&\makecell{Shore based\\(RGB \& IR)}&--&--&--&20&Yes: \href{http://www.diag.uniroma1.it//~labrococo/MAR/}{\textcolor{green}{Click Here}}\\
		\hline
		
	\textbf{Botlek} \cite{ghahremani2017self}&\makecell{Maritime\\(Vessel Detection)}&&\checkmark&\makecell{(RGB)}&$1536\times2048$&--&--&$>$48K&No\\
		\hline
		
	\textbf{MSD} \cite{chen2021improved}&\makecell{Maritime\\(Ship Detection)}&&\checkmark&\makecell{Satellite based\\(panchromatic)}&$1000\times1000$&4&--&1015&No\\
		\hline
		
	\textbf{MODD} \cite{kristan2015fast}&\makecell{Maritime\\(Obstacle Detection)}&\checkmark&&\makecell{USV based\\(RGB)}&$640\times480$&2&--&\makecell{12\\(4454 fully annotated frames)}&Yes: \href{https://www.vicos.si/resources/modd/}{\textcolor{green}{Click Here}}\\
		\hline
		
	\textbf{IPATCH} \cite{patino2016pets}&\makecell{Maritime\\(Auto Protection)}&\checkmark&&\makecell{On-board\\(Visual \& IR)}&\makecell{$640\times480$\\$640\times512$}&--&--&14&Yes\\
		\hline
		
	\textbf{FGSD} \cite{chen2020fgsd}&\makecell{Maritime\\(Ship Detection)}&&\checkmark&\makecell{Satellite based\\(RGB)}&$930\times930$&43&5634 &\makecell{4736\\2612  annotated}&Yes: \textcolor{green}{Coming Soon} \\
		\hline
		
	\textbf{ShipRSImageNet} \cite{zhang2021shiprsimagenet}&\makecell{Maritime\\(Ship Detection)}&&\checkmark&\makecell{Satellite based\\(RGB)}&$930\times930$&50&17573&$>$3435&Yes: \href{https://github.com/zzndream/ShipRSImageNet}{\textcolor{green}{Click Here}}\\
		\hline
	\textbf{BCCT200} \cite{rainey2011object}&\makecell{Maritime\\(Ship Detection)}&&\checkmark&\makecell{Satellite based\\(GS)}&NF&4&--&800&Yes\\
		\hline	
	
	\textbf{Chen~\textit{et al.}} \cite{chen2020video}&\makecell{Maritime\\(Ship Detection)}&\checkmark&&\makecell{UAV based\\(RGB)}&$720\times480$&--&--&\makecell{2\\(3000 frames)}&Yes: \textcolor{green}{Under Request}\\
		\hline
	\textbf{Airbus Ship Detection}&\makecell{Maritime\\(Ship Detection)}&&\checkmark&\makecell{Satellite based\\(RGB)}&$768\times768$&--&--&\makecell{$>$192K}&Yes: \href{https://www.kaggle.com/c/airbus-ship-detection/data}{\textcolor{green}{Click Here}}\\
		\hline
	
	\textbf{SeaDronesSees} \cite{varga2022seadronessee} &\makecell{Maritime\\(Search and Rescue)}&\checkmark&\checkmark&\makecell{UAV based\\(RGB \& NIR \& RE)}&\makecell{$3840\times2160$\\$5456\times3632$}&6&400K&\makecell{5630 images, 208 short videos,\\
	22 videos\\($>$393K and 54K frames)}&Yes: \href{https://seadronessee.cs.uni-tuebingen.de./}{\textcolor{green}{Click Here}}\\
		\hline
		
	\textbf{MOBDrone} \cite{cafarelli2022mobdrone} &\makecell{Maritime\\(Search and Rescue)}&\checkmark&&\makecell{UAV based\\(RGB)}&\makecell{--}&5&$>$180K&\makecell{66\\(126170 annotated frames)}&Yes: \href{http://aimh.isti.cnr.it/dataset/MOBDrone/}{\textcolor{green}{Click Here}}\\
		\hline		
	\end{tabular}}
\end{table*}
\textbf{Singapore Maritime Dataset (SMD)} \cite{prasad2017video} was collected using Canon 70D cameras around Singapore waters. The dataset comprises on-shore and on-board videos as well as Near Infra Red (NIR) videos.\\
\textbf{MarDCT} \cite{Bl-Io-Pe-15} is comprised of visible and infrared images captured mostly from buildings near congested marine routes in Italy.\\
\textbf{Botlek Dataset} \cite{ghahremani2017self} is a dataset containing an image set sampled from video recordings (6 view points) from the Botlek region in the port of Rotterdam, Netherlands. The dataset captures a variety of weather conditions, object sizes, camera positions, occlusions, \textit{etc.}\\
\textbf{MSD} \cite{chen2021improved} or Multi-class Ship Dataset was constructed to classify ships into four different classes: big ships, middle ships, small ships and moving ships. The images were collected from GF-1 and GF-2 satellites, which covered different landscapes, light and weather conditions.\\
\textbf{MODD dataset} \cite{kristan2015fast} is a Marine Obstacle Detection Dataset especially designed for the identification of small or large obstacles in maritime environments. This dataset contains 12 video sequences captured by Unmanned Surface Vehicles (USV). The videos were recorded from multiple platforms, often by a small 2.2 meter USV. \\
\textbf{IPATCH Dataset} \cite{patino2016pets} contains 14 multisensor observations (visible and thermal) from the coast of Brest, France. The purpose of this dataset is to protect merchant ships from piracy. \\
\textbf{Fine-Grained Ship Detection (FGSD)} \cite{chen2020fgsd} is a dataset with high resolution remote sensing images acquired from a Google Earth platform. This includes ship instances from 17 different ports (USA, China, Spain, Japan) around the world. The resolution  of the images ranged from  0.12m  to 1.93m.   \\
\textbf{ShipRSImageNet} \cite{zhang2021shiprsimagenet} is amongst the largest remotely sensed image datasets for fine-grained ship classifications which includes diverse complex environments and small ships. This makes this dataset suitable for deep learning-based methods. This dataset consists of images compiled  from a variety of sensor platforms and other datasets, in particular xView, HRSC2016, FGSD, \textit{etc.} \\
\textbf{BCCT200} \cite{rainey2011object} is one of the earliest vessel detection datasets consisting of different ship types of barges, cargoes, containers, and tankers.  \\
\textbf{Chen~\textit{et al.} Dataset} \cite{chen2020video} contains several maritime videos acquired from coastal areas near Shanghai in China. Videos were acquired under two scenarios: straight-forward and irregular movements. \\
\textbf{Airbus Ship Detection} is a dataset and Kaggle competition to benchmark methods for localizing ships in remote sensing images. \\
\textbf{SeaDronesSees Dataset} \cite{varga2022seadronessee} is the first large-scale annotated UAV-based dataset of swimmers in open waters. The class labels are swimmers, floaters (swimmers with life jackets), life jackets, swimmers (person on boat not wearing a life jacket), floaters (person on boat wearing a life jacket), and boats. The dataset offers three challenges: object detection, single-object tracking, and multi-object tracking. \\
\textbf{MOBDrone} \cite{cafarelli2022mobdrone} is a large-scale dataset captured by a UAV in the Gombo beach of the Migliarino, Pisa, Italy at a height of 10-60 meters. A total of 5 types of objects are included in the dataset: people, boats, woods, life buoys, and surface boards.\\

\subsection{Evaluation Metrics}
\begin{figure*}[]
	\centering
	\begin{adjustbox}{width = 0.95\linewidth}
		\begin{tabular}{ccc}
			\includegraphics[scale=0.5]{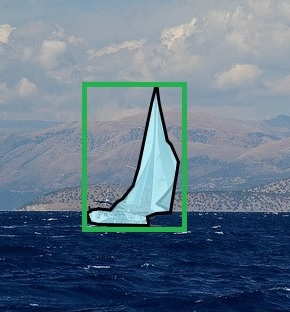} & 
			\includegraphics[scale=0.225]{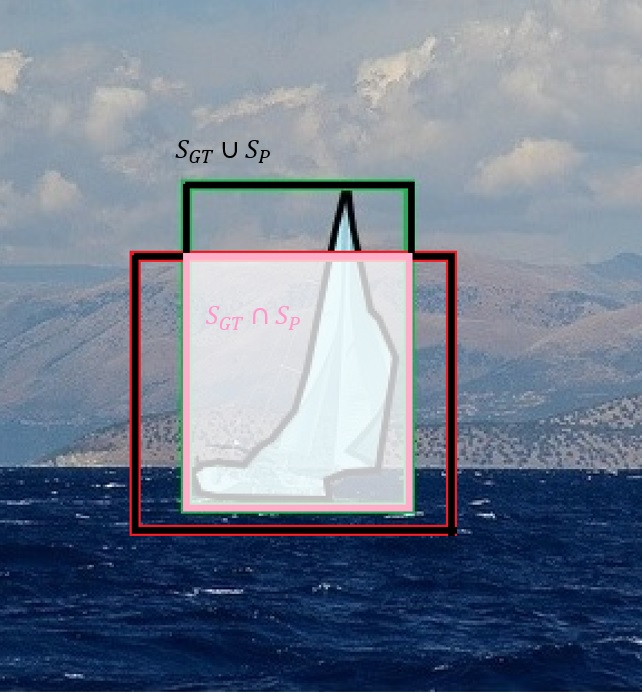} &
			\includegraphics[scale=0.225]{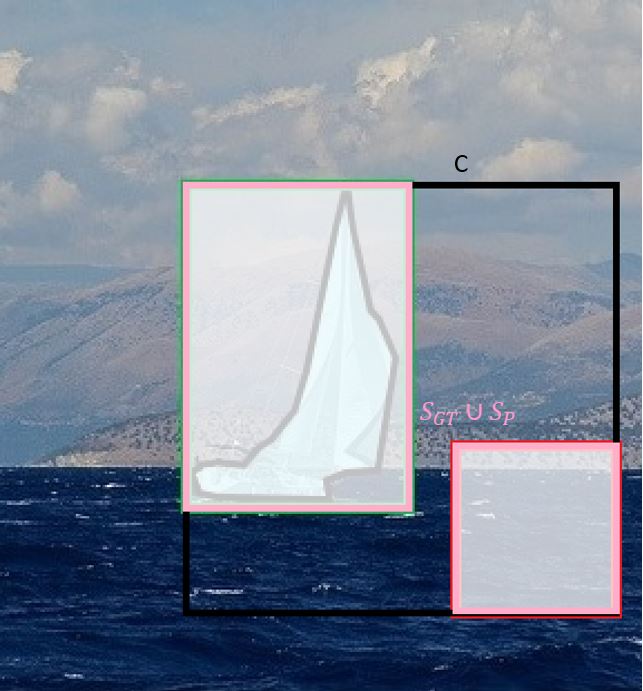}\\
				(a) & (b) & (c)\\
			\includegraphics[scale=0.18]{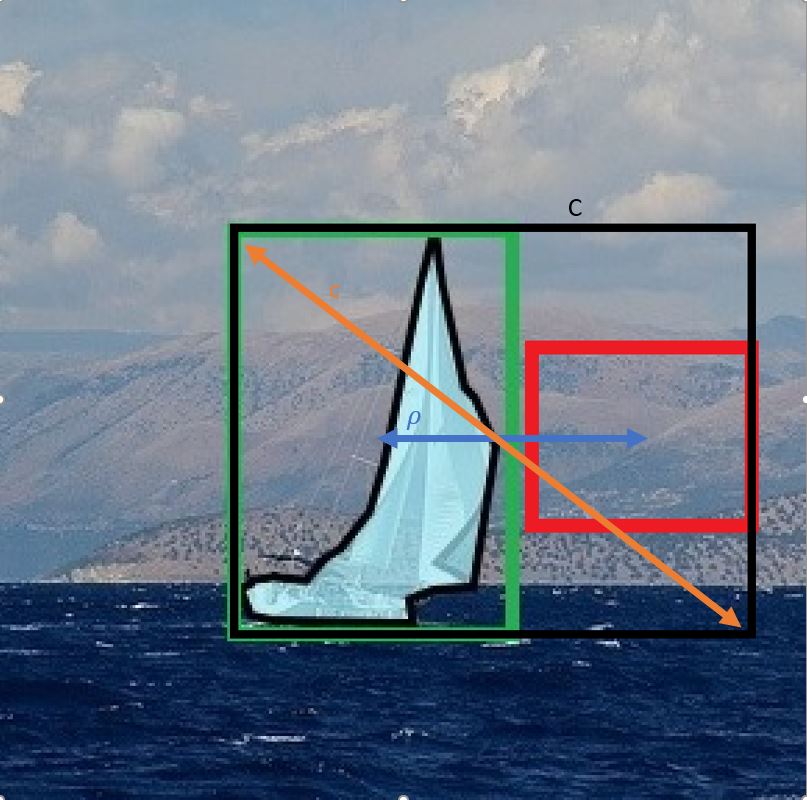}&
			\includegraphics[scale=0.18]{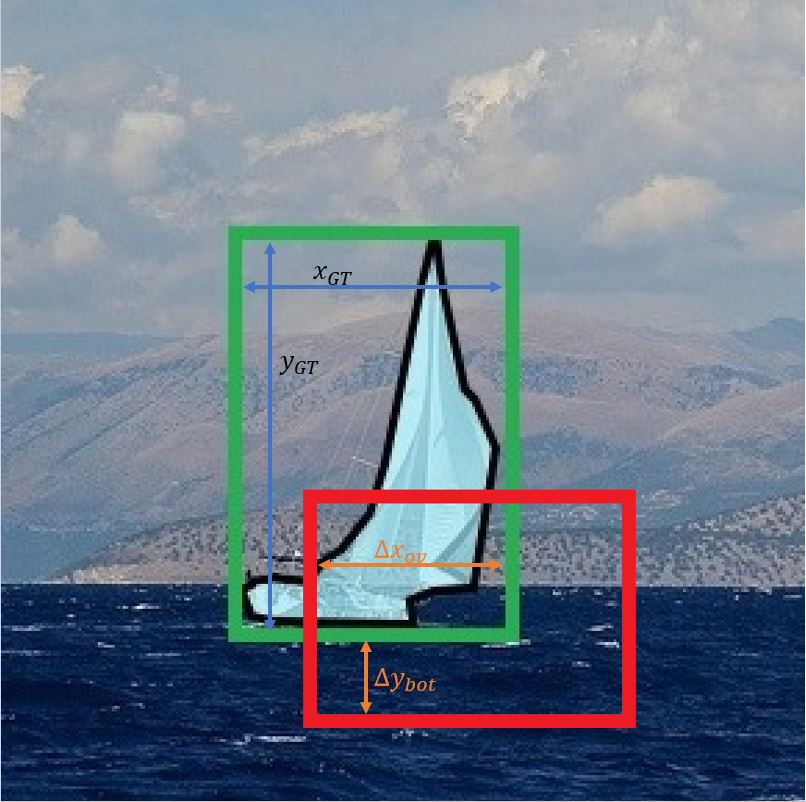}&
			\includegraphics[scale=0.18]{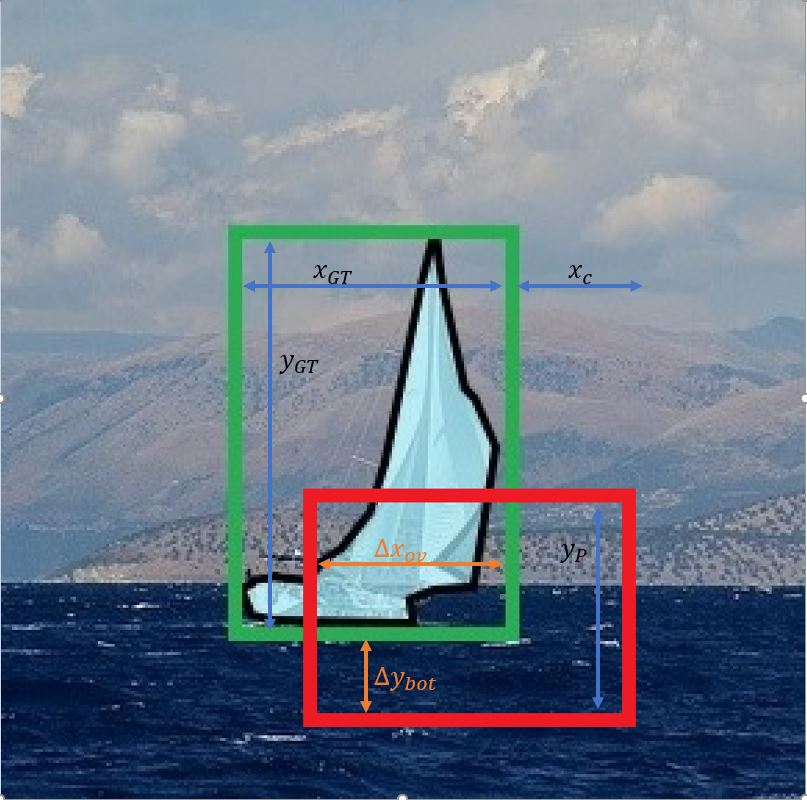}\\
		  (d) & (e)& (f)\\
		\end{tabular}
	\end{adjustbox}
	\caption{Parameters of evaluation metrics introduced in Sec. 6.2 for predicted red box, (a) GT, (b) parameters for IoU, (c) parameters for GIoU, (d) parameters for CIoU, (e) parameters for BEP1 and (f) parameters for BEP2.  }
	\label{Eval}
\end{figure*}
\subsubsection{General Measures}
\textbf{Intersection over Union \cite{everingham2010pascal}:} Since the output of an object detection method and its corresponding ground truth are the coordinates of bounding boxes, the Intersection over Union (IoU) is used to quantify the similarity between the areas of these two bounding boxes; Ground Truth (GT) and Predicted (P). when the bounding boxes are indexing the same pixels, this measure is expected to return a value one in the best case, and zero in the worst case when the boxes are not overlapped at all. Using the set notations, the IOU is given by
\begin{equation}
	IoU=\frac{|S_{GT}\cap S_P|}{|S_{GT}\cup S_P|},
\end{equation}
where $S$ indicates the pixels as a set, $|.|$ is the size of a set, $\cap$ and $\cup$ are the intersection and union, respectively. Fig. \ref{Eval}(a) shows the GT in green and in Fig. \ref{Eval}(b) the intersection (pink square) and union (black boundaries) are clearly shown in the image assuming the red bounding box as the prediction.\\
\textbf{Precision, Recall and Accuracy:} These are well known measures in classification tasks defined for categorical outputs. Object detection, however, uses bounding boxes whose similarity is shown through continuous numbers ranging from 0 to 1.  A threshold is therefore applied to the IoU in order to use such measures for object detection. Predicted bounding boxes are accepted as true positives (accurate recovery of the ground truth bounding box) if the corresponding IoUs exceed the threshold, otherwise they are considered false positives. Specifically, precision is defined as the number of correctly detected bounding boxes compared to the total number of detected or predicted boxes. Recall, on the other hand, is defined as the number of correctly detected bounding boxes over the total number of ground truth boxes. It is therefore necessary to make a trade-off between recall and precision. Finally, accuracy is defined as the total number of correctly labeled bounding boxes (either positive or negative) over the total number of evaluated boxes. \\
\textbf{Average Precision (AP):} The trade-off between Precision (Pr) and Recall (Re) prevents comparing two given methods using a single precision value for a fixed recall. Rather, precision needs to be on average better across all recall values. Therefore, the precision-recall curve can be drawn for each class label and the area under the curve can be determined. A method is better if its computed area is larger than that of its competitors. Precisely, the AP is given by:
\begin{equation}
	AP=\int_{0}^{1}Pr(Re)dRe,
\end{equation}
where $Pr(Re)$ indicates the dependence of precision on the recall value.
\\
\textbf{mean Average Precision (mAP) and mean Average Recall (mAR):} Due to the fact that AP is defined over a single class label, it is not universal across all classes. In order to generalize this measure, the mAP computes the average over all the classes. In other words, for mAP we have
\begin{equation}
	mAP=\frac{1}{C}\sum_{i=1}^{C}AP_i,
\end{equation}
where $C$ denotes the number of classes. Observe that the average above is computed based on a single predefined threshold, \textit{e.g.}, 0.5. In a broader sense, this average can be computed in terms of different threshold values, notably from 0.5 to 0.95 with a 0.05 step size. This particular setup is denoted as $mAP^{@[0.5,0.95]}$ in \cite{lin2014microsoft}. Similarly we have the same concept for recall, with the equivalent metric being mAR which is defined for the average of the individual recalls over the number of classes. \\
\textbf{Frame Per Second (FPS):} In addition to the measures which evaluate the ability of the detection methods in recovering the true objects, FPS measures the running time of these techniques to evaluate their applicability to video or real time detection. The higher FPS implies that the method is faster and can potentially be applied to real-time video-based small object detection. \\
\textbf{Degrade of Reduction (DOR) \cite{chen2020survey}:} This measure indicates the performance gap between the AP of medium/large objects and that of small objects. SOD performance is weaker when DOR is larger. \\
\textbf{FPPI:} The average number of false positives per image when recall is 0.5 and the recall when FPPI is 1 are two other measures that have been used for evaluation of SOD methods \cite{rozantsev2016detecting}. Ideally, we aim for smaller FPPI and higher recall for a fixed FPPI. \\
\textbf{Intersection over Detection (IoD):} This measure is similar to IoU with a minor change in the denominator. In other words, the IoD is given by:
\begin{equation}
	IoD=\frac{|S_{GT}\cap S_P|}{|S_P|}.
\end{equation}
As a result of this change, small objects won't be missed in applications where accurate detection of true objects is crucial at the cost of more false positives.
\\
\textbf{Generalized IoU (GIoU) \cite{rezatofighi2019generalized}:} If two boxes are not overlapping, IoU is not helpful during the learning process since it is always zero no matter how distinct the boxes are. For this reason, the GIoU loss has been proposed as a solution to Gradient vanishing. Thus the GIoU is given by:
\begin{equation}
	GIoU=IoU-\frac{|C\backslash S_{GT}\cup S_P|}{|C|},
\end{equation}
where $C$ is the smallest box containing both GT and P bounding boxes and ``$\backslash$" means excluding the set in the right from the left set. Fig. \ref{Eval}(c) shows an example of the use of this metric ($C$ and $S_{GT}\cup S_P$).
\\
\textbf{Complete IoU (CIoU)} \cite{zheng2021enhancing}: As a result of its inability to exploit geometrical factors in the metric, GIoU suffers from slow convergence and inaccurate regression. In contrast, CIoU improves performance by considering three main geometrical factors, namely the overlaped area, the distance and the aspect ratio to improve the performance. It is given by: 
\begin{equation}
	CIoU=IoU-\frac{\rho^2(GT,P)}{c^2}-\alpha V,
\end{equation}
where $\rho$ is the Euclidean distance between the central points of the boxes, $c$ is the diagonal length of the smallest box containing both GT and P bounding boxes, $\alpha$ is the trade-off parameter, and finally $V$ is the consistency of aspect ratios. Fig. \ref{Eval}(d) shows an example of the use of this metric ($\rho$ and $c$).  \\
\textbf{Miss Rate (MR)}: Even though the trade-off between false positives and miss detection rate matters in most applications, in some real world problems (\textit{e.g.}, pedestrian and tumor detection) the MR is the main objective since the object should not be missed in order to avoid major consequences (\textit{e.g.}, accident or cancer). A smaller MR is always desirable \cite{dollar2009pedestrian}.\\ 
\textbf{Error Rate (ER):} Deep network training can also be optimized by minimizing a measure of error. In this case, the ER is defined as the total number of miss classified pixels over the total number of pixels.\\
\textbf{Normalized Wasserstein Distance (NWD)}\cite{wang2021normalized}: As opposed to the aforementioned metrics, which treat bounding boxes as deterministic variables, here the bounding boxes are represented by multivariate Gaussian densities. The similarity is then calculated by an exponential function of the existing Optimal Transport (OT) theory (\textit{i.e.}, Wasserstein distance). The benefit of this approach lies in assigning different weights to different pixels, putting more emphasis on the central pixels. In other words, the similarity is given by
\begin{equation}
	NWD(GT,P)=\exp\{-\frac{\sqrt{W_2^2(GT,P)}}{c}\},
\end{equation}
where $c$ is a learnable constant, and $W_2^2(GT,P)=\|\textbf{m}_1-\textbf{m}_2\|_2^2+\|\boldsymbol{\Sigma}_1^{1/2}-\boldsymbol{\Sigma}_2^{1/2}\|_F^2$ is the Wasserstein distance between two ground truth and predicted bounding boxes where $\textbf{m}$ is the centre of the boxes and $\boldsymbol{\Sigma}$ is their covariance. 
\subsubsection{Specific to Maritime}
\textbf{Intersection over Ground truth (IoG) \cite{prasad2018object}: }As with autonomous driving, detecting ships in maritime environments is very important in order to avoid collisions. The bounding boxes in maritime SOD tend to be wider than those in other applications because of wakes and waves. False positives are caused by using the standard IoU metric. The modified metric IoG can help mitigate this issue and is defined by:
\begin{equation}
	IoG=\frac{|S_{GT}\cap S_P|}{|S_G|}.
\end{equation}
\\
\textbf{Bottom Edge Proximity 1 (BEP1) \cite{prasad2018object}:} Objects in the sea may be characterized by a solid dense hull having a larger possibility of detection and a sparse mast region. The standard IoU criteria may regard the detected object as a false alarm since the ground truth covers both dense hull and mast regions. The BEP1 metric helps to avoid such inaccuracies and it is given by:
\begin{equation}
	BEP1=X(1-Y); X=\frac{\Delta x_{ov}}{ x_{GT}}, Y=\frac{\Delta y_{bot}}{y_{GT}}. \nonumber
\end{equation}
The parameters for this metric are as shown in Fig. \ref{Eval}(e).\\
\textbf{Bottom Edge Proximity 2 (BEP2) \cite{prasad2019object}:} BEP2 is symmetric with respect to ground truth and predicted bounding boxes while BEP1 is biased toward ground truth. The BEP2 is defined as
\begin{equation}
	BEP2=X(1-Y); X=\frac{\Delta x_{ov}}{ x_{GT}+x_c}, Y=\frac{\Delta y_{bot}}{min(y_{GT},y_P)}.\nonumber  
\end{equation}
The parameters for this metric are as shown in Fig. \ref{Eval}(f).\\

\subsection{Performance Evaluation}
In this section, we assess the performance of the discussed SOD methods on different large-scale datasets. For the generic SOD evaluation, we selected the popular image datasets: Tsinghua-Tencent 100K and MS COCO. For the analysis of video-based techniques, we selected the USC-GRAD-STDdb and UAVDT, which are relatively challenging. This paper uses all performance measures taken from the original papers, or their websites. Research usually compares methods using a subset of these datasets (for example, MS COCO) since some of these datasets are not specifically designed for SOD. The table captions clearly indicate the setups corresponding to the reported results. \\
SOD datasets for maritime applications are still rare, so most papers perform performance analyses on datasets that they have designed themselves. As a result, the maritime case study results were presented together with the generic methods using four image datasets, including TinyPerson, SeeDronesSees, WSODD, and ShipRSImageNet. For video datasets, we selected Seagull and SMD since they are more popular.\\
Tables \ref{tab4} to \ref{tab7} show the results for generic small objects and similarly, Tables \ref{tinyperson} and \ref{tab9} show the results for maritime small objects.    
\subsubsection{Generic SOD Performance Results}
\par\noindent\textbf{Tsinghua-Tencent 100K.} Table~\ref{tsinghua} reports the detection performance of the state-of-the-art methods on images with small objects, whose number of pixels are in the range of (0,32], in terms of recall, accuracy and F1-score. As shown in Table~\ref{tsinghua}, Liang~\textit{et al.}\cite{liang2018small} achieved the best Recall of 93.0\% and a moderate accuracy of 84.0\%. In contrast, YOLOv3-Final~\cite{wan2021efficient} attained the best accuracy of 91.0\% with a recall of 91.0\%, leading to the best F1-score of 91.0\%.  
\par\noindent\textbf{MS COCO.}
Table \ref{tab5} shows the detection results of deep learning-based methods on MS COCO dataset. For comparison, we report $\text{mAP}^{@0.5}$ and $\text{mAP}^{@[0.5,0.95]}$. Since the comparison was made using different setups, we denote the results of object detection with sizes smaller than $32 \times 32$ with normal values, the results of objects with sizes smaller than $16 \times 16$  with ``+" and the results on a subset of MS COCO including the three classes of stop signs, mice, and fire hydrants with values marked with ``*". As shown, Full Deformable DETR~(arXiv20)\cite{zhu2020deformable} achieved the best $\text{mAP}^{@[0.5,0.95]}=34.4$. In general, smaller objects produce poorer results. FPN~(CVPR17)\cite{lin2017feature} achieves the best values for both $\text{mAP}^{@0.5}$ and $\text{mAP}^{@[0.5,0.95]}$ for smaller objects, with values of 11.8 and 4.8, respectively. Finally DETR-GQPos-SiA~(arXiv21)\cite{jiang2021guiding} achieves the best $\text{mAP}^{@0.5}$ of 24.4 for normal small objects on MS COCO. Table \ref{tab5} also shows the results for the MS COCO subset separately. As can be observed, transformer-based deep learning methods currently have the best SOTA results.   
\begin{table}[]
\caption{Detection performance (\%) for small-scale objects on Tsinghua-Tencent 100K~\cite{zhu2016traffic}}
\label{tsinghua}
\centering
\label{tab4}
\begin{tabular}{lccc}
\toprule
            & Recall & Accuracy & F1-score\\ \hline
Fast RCNN~(ICCV15)\cite{girshick2015fast}& 46.0 &74.0  & 56.7       \\
Faster RCNN~(NIPS15)\cite{ren2015faster} &  49.8&24.1   &32.5       \\
SSD~(ECCV16)\cite{liu2016ssd} &  43.4&25.3&32.0      \\
Zhu~\textit{et al.}~(CVPR16)\cite{zhu2016traffic} &87.4&81.7&84.5\\
FPN~(CVPR17)\cite{lin2017feature} &  78.6&77.3&77.9    \\
Perceptual GAN~(CVPR17)\cite{li2017perceptual}&89.0&84.0&86.4 \\
Pon~\textit{et al.}~(CRV18)\cite{pon2018hierarchical} &65.0&24.0&35.1\\
Liang~\textit{et al.}~(PCM18)\cite{liang2018small}&\textbf{93.0}&84.0&88.3\\
Song~\textit{et al.}~(JSA19)\cite{song2019efficient} &88.0&85.0&86.5 \\
Noh~\textit{et al.}~(ICCV19)\cite{noh2019better} &92.6&84.9&88.6 \\
MR-CNN~(ACCESS19)\cite{liu2019mr}&89.3&82.9&86.0\\
Wang~\textit{et al.}~(ITS20)\cite{wang2020traffic}&89.4&87.3&88.3\\
YOLOv3-Final~(JSPS21)\cite{wan2021efficient} &91.0&\textbf{91.0}&\textbf{91.0} \\
SODNet~(RS22)\cite{qi2022small}&90.0&85.5&87.7\\
Min~\textit{et al.}~(ITS22)\cite{min2022traffic} &92.3&88.1&90.2 \\
\bottomrule
\end{tabular}

\end{table}

\begin{table}[]
\caption{Detection performance (\%) for small-scale objects on MS COCO image dataset. ``${*}$" indicates average over just three classes of stop sign, mouse, fire hydrant. Currently, the leadership for small objects on MS COCO dataset belongs to Noah CV Lab (Huawei) with $\text{mAP}^{@[0.5,0.95]}=40.7$. The results are by default for objects smaller than $32\times32$ pixels. ``$+$" indicates that the results are for object sizes smaller than $16\times16$.}
\centering
\label{tab5}
\scalebox{0.88}{\begin{tabular}{lccc}
\toprule
 &$\text{mAP}^{@0.5}$  $\uparrow$  & $\text{mAP}^{@[0.5,0.95]}$  $\uparrow$  \\ \hline
Faster R-CNN~(NIPS2015)\cite{ren2015faster}& $5^+$ &$15.6$, $1.5^+$ \\ 
  
Faster R-CNN+FPN~(NIPS2015)\cite{ren2015faster}& -- &27.2 \\
  
R-FCN~(NIPS16)\cite{dai2016r}&--&10.8\\
 
SSD~(ECCV16)\cite{liu2016ssd}&-- &10.9 \\
 
FPN~(CVPR17)\cite{lin2017feature} & $\textbf{11.8}^+$ &18.2, $\textbf{4.8}^+$\\
 
RetinaNet~(ICCV17)\cite{lin2017focal} & $9.1^+$& 21.8, $4.5^+$ \\

RFBNet~(ECCV18)\cite{liu2018receptive}& 16.2 &  --\\
 
YOLOv3~(arXiv18)\cite{redmon2018yolov3}&--&18.3\\
 
SOD-MTGAN~(ECCV18)\cite{bai2018sod}& --&25.1 \\
 
Noh~\textit{et al.}(ICCV19)~\cite{noh2019better}&-- &16.2\\

Kisantal~\textit{et al.} (arXiv19)~\cite{kisantal2019augmentation}& -- & 17.9  \\

FCOS~(ICCV19)\cite{tian2019fcos}&--&24.4\\

SSD-MSN~(IEEE ACCESS19)\cite{chen2019ssd}&-- &29.4  \\

FSAF~(CVPR19)\cite{zhu2019feature}&-- &29.7  \\

DR-CNN sum~(AI20)\cite{liu2020small}& 18.3 & -- \\

DR-CNN concat.~(AI20)\cite{liu2020small}& 18.6 & -- \\

ViT-FRCNN~(arXiv20)\cite{beal2020toward}&-- &17.8 \\

DETR~(ECCV20)\cite{carion2020end}&-- &21.9 \\

DETR-DC5&-- & 23.7\\

Deformable DETR~(arXiv20)\cite{zhu2020deformable}& -- & 26.4\\

Two Stage Deformable DETR~(arXiv20)\cite{zhu2020deformable}& -- & 28.8\\

Full Deformable DETR~(arXiv20)\cite{zhu2020deformable}& -- &$\textbf{34.4}$ \\

ATSS~(CVPR20)\cite{zhang2020bridging} & --&33.2\\

YOLOv5s \cite{jocher2020yolov5}& --& 18.8 \\

TSD~(CVPR20) \cite{song2020revisiting}& --&33.8\\

STDnet-C3~(EAAI20)\cite{bosquet2020stdnet}&$11.4^+$&$5.5^+$\\

YOLOS~(NIPS21)\cite{fang2021you}&--&19.5\\

UP-DETR~(CVPR21)\cite{dai2021up}&--&20.8\\

SOF-DETR\cite{dubey2021improving}& --& 21.7 \\

ViDT w.o. Neck~(arXiv21)\cite{song2021vidt}& --& 21.9 \\

ViDT~(arXiv21)\cite{song2021vidt}& --& 30.6 \\

SMCA~(ICCV21)\cite{gao2021fast}&22.8&--\\

DETR-GQPos~(arXiv21)\cite{jiang2021guiding}&23.1&--  \\

DETR-GQPos-SiA~(arXiv21)\cite{jiang2021guiding}&$\textbf{24.4}$&--\\

FP-DETR~(ICLR22)\cite{wang2021fp}&--&27.5\\

SODNet~(RS22)\cite{qi2022small}&-- &20.1 \\

RFSOD~(RTIP22)\cite{amudhan2021rfsod}        &$\textbf{59.09}^{*}$ & --\\
   
RFSODTL~(RTIP22)\cite{amudhan2021rfsod} & $56.42^{*}$        &-- \\

QueryDet~(CVPR22)\cite{yang2022querydet}&--&25.24\\

RESC~(NCA22)\cite{wang2022resc}&--&26.2\\
   
$\text{D}^2$ETR~(arXiv22)\cite{lin2022d} & --       &22 \\

Deformable $\text{D}^2$ETR~(arXiv22)\cite{lin2022d} & --       &31.7 \\
\bottomrule
\end{tabular}}

\end{table}
\begin{table}[]
\caption{Detection performance (\%) for small-scale objects on USC-GRAD-STDdb video dataset \cite{bosquet2020stdnet}. +k indicates that the anchors were defined by the k-means algorithm and the ``$*$" indicates that they were run on Caffe2 framework. The results are by default for the objects smaller than $16\times16$ pixels. }
\centering
\label{tab6}
\scalebox{0.75}{\begin{tabular}{lcccc}
\toprule
            & mAP$^{@0.5}$  $\uparrow$ & $\text{mAP}^{@[0.5,0.95]}$  $\uparrow$ & FPPI  $\downarrow$ & FPS  $\uparrow$\\ \hline
Faster R-CNN~(NIPS15)+k\cite{ren2015faster} & $44$&$14.4$ &0.95&2.6\\

FPN~(CVPR17)\cite{lin2017feature} & $50.8$&$16.3$&$0.29$&$3$\\

FPN~(CVPR17)+k\cite{lin2017feature} & $50.7$&$16.8$&$0.31$&$3.5$\\

RetinaNet~(ICCV17)\cite{lin2017focal} & $47.6$&$16.2$&$0.47$&$\textbf{6.5}^{*}$\\

FGFA~(ICCV17)\cite{zhu2017flow}&$37.5$&$11.7$&--&--\\

Cascade-FPN~(CVPR18)\cite{cai2018cascade}&$55.9$&$17.4$&--&--\\

RDN~(ICCV19)\cite{deng2019relation}&$48.6$&$15.5$&--&--\\

FANet(short term)~(arXiv20)\cite{cores2020spatio}& $48.5$ & $17.6$&--  &--      \\

FANet(short\&long term)~(arXiv20)\cite{cores2020spatio}& $49.9$ & $18.3$ &--  &--      \\

MEGA~(CVPR20)\cite{chen2020memory}&$53.1$&$17.4$&--&--\\

STDnet-C3~(EAAI20)\cite{bosquet2020stdnet}& $57.4$ & $20$ &0.22 & 3.7       \\

STDnet-bST~(EAAI20)\cite{bosquet2020stdnet}& $59.7$ & $20.6$ &\textbf{0.2}  &--      \\

STDnet-ST~(PR21)\cite{bosquet2021stdnet}& $62.1$ & $20.1$ &--  &--      \\

STDnet-ST++~(PR21)\cite{bosquet2021stdnet}& $\textbf{63.4}$ & $\textbf{21.4}$ &--  &--      \\

\bottomrule
\end{tabular}}

\end{table}

\begin{table}[]
\caption{Detection performance (\%) for small-scale objects on UAVDT video dataset \cite{du2018unmanned}. The results are by default for objects smaller than $32\times32$ pixels. ``$+$" indicates the results of object sizes smaller than $16\times16$.}
\label{tab7}
\centering
\scalebox{0.9}{\begin{tabular}{lcccc}
\toprule
            &mAP$^{@0.5}$  $\uparrow$& $\text{mAP}^{@[0.5,0.95]}$  $\uparrow$ \\ \hline
Faster R-CNN+FPN~(ECCV18)\cite{du2018unmanned} &$26^{+}$ &8.1\\
R-FCN~(NIPS16) \cite{dai2016r}&$32.5^+$& 4.4\\

SSD~(ECCV16)\cite{liu2016ssd} &$23.5^+$& 7.1\\

RON~(CVPR17)\cite{kong2017ron}&$19.7^+$&2.9\\

FPN~(CVPR17)\cite{lin2017feature} &$29.7^+$&$11.8^+$\\

FGFA~(ICCV17)\cite{zhu2017flow}&$20.7^+$&$6.3^+$\\

Cascade-FPN~(CVPR18)\cite{cai2018cascade}&$30.5^+$&$12^+$\\

RDN~(ICCV19)\cite{deng2019relation}&$27.9^+$&$9.3^+$\\

ClusDet~(ICCV19)\cite{yang2019clustered}&--&9.1\\

YOLOv5s \cite{jocher2020yolov5}&--&9.8\\

MEGA~(CVPR20)\cite{chen2020memory}&$26.6^+$&$9.2^+$\\

STDnet++~(EAAI20)\cite{bosquet2020stdnet} &$35.4^+$ &$12.6^+$     \\

STDnet-ST++~(PR21)\cite{bosquet2021stdnet}& $\textbf{36.4}^+$ & $\textbf{13.3}^+$      \\

SODNet~(RS22)\cite{qi2022small}&--&11.9\\
\bottomrule
\end{tabular}}

\end{table}

\par\noindent\textbf{USC-GRAD-STDdb.}
For the evaluation on video sequences, we selected the recently released dataset, USC-GRAD-STDdb to compare existing SOTA methods. Table \ref{tab6} shows the results obtained on this dataset for various metrics of $\text{mAP}^{@0.5}$, $\text{mAP}^{@[0.5,0.95]}$, FPPI, and FPS. By default, the results for this particular dataset are reported for sizes smaller than $16\times16$. The team who collected the USC-GRAD-STDdb dataset proposed STDnet-ST++ (PR21)\cite{bosquet2021stdnet}, which remains the leading technique in terms of average precision. In terms of FPPI, STDnet-bST (EAAI20)\cite{bosquet2020stdnet}, another framework proposed by the same team, performs best. Finally, RetinaNet~(ICCV17)\cite{lin2017focal} achieves the best results in terms of runtime speed.  
\par\noindent\textbf{UAVDT.}
As for this video dataset, Table \ref{tab7} shows the results for $\text{mAP}^{@0.5}$, and $\text{mAP}^{@[0.5,0.95]}$. The values are by default for objects smaller than $32\times32$. However, smaller sizes than $16\times16$ are indicated by ``+". As for USC-GRAD-STDdb dataset, STDnet-ST++~(PR21)\cite{bosquet2021stdnet} is seen again to be the leading method for generic small object detection task. \\  
\subsubsection{Maritime SOD Performance Results}


\begin{table*}[]
\caption{Detection performance (\%) for small-scale objects on TinyPerson~\cite{yu2020scale}. MR and AP denote Miss Rate and Average Precision. The superscripts of MR and AP denote the size splits, where ``tiny" refers to the size range [2,20] and ``small" refers to the size range [20,32]. The subscripts of MR and AP denote the IOU thresholds used for the evaluation.}
\centering
\label{tinyperson}
	\scalebox{0.9}{\begin{tabular}{lcccccccc}
\toprule
            & $MR^{tiny}_{50}$ $\downarrow$ & $MR^{small}_{50}$ $\downarrow$& $MR^{tiny}_{25}$ $\downarrow$& $MR^{tiny}_{75}$ $\downarrow$& $AP^{tiny}_{50}$ $\uparrow$& $AP^{small}_{50}$ $\uparrow$& $AP^{tiny}_{25}$ $\uparrow$& $AP^{tiny}_{75}$ $\uparrow$ \\ \midrule
Faster RCNN-FPN~(CVPR17)\cite{lin2017feature}&87.78&71.31&77.35&98.40&43.55&56.69&64.07&5.35\\
RetinaNet~(ICCV17)\cite{lin2017focal}&92.40&81.75&81.56&99.11&30.82&43.38&57.33&2.64\\
DSFD~(CVPR19)\cite{li2019dsfd}&93.47&78.72&78.02&99.48&31.15&51.64&59.58&1.99\\
Adaptive FreeAnchor~(NIPS19)\cite{zhang2019freeanchor}&88.97&73.67&77.62&98.70&41.36&53.36&63.73&4.00\\
FCOS~(ICCV19)\cite{tian2019fcos}&96.12&84.14&89.56&99.56&16.9&35.75&40.49&1.45 \\
Libra RCNN~(CVPR19)\cite{pang2019libra}&89.22&74.86&82.44&98.39&44.68&62.65&64.77&6.26\\
Grid RCNN~(CVPR19)\cite{lu2019grid}&87.96&73.16&78.27&\textbf{98.21}&47.14&62.48&68.89&6.38\\
RetinaNet-SM~(WACV20)\cite{yu2020scale}&88.87&71.82&77.88&98.57&48.48&63.01&69.41&5.83\\
Faster RCNN-FPN+MSM~(WACV20)\cite{yu2020scale}&\textbf{85.86}&\textbf{68.76}&74.33&98.23&50.89&65.76&71.28&6.66 \\
RetinaNet+SM with S-$\alpha$~(WACV21)~\cite{gong2021effective}&87.00&69.25&74.72&98.41&52.56&65.69&\textbf{73.09}&6.64\\
Faster RCNN-FPN+MSM with S-$\alpha$~(WACV21)~\cite{gong2021effective}&86.18&69.28&\textbf{73.90}&98.24&51.41&65.97&72.25&6.69\\
Faster RCNN-FPN-MSM+~(ICASSP21)\cite{jiang2021sm+}&--&--&--&--&\textbf{52.61}&\textbf{67.37}&72.54&\textbf{6.72}\\
\bottomrule
\end{tabular}}

\end{table*}

\begin{table*}
	\caption{Detection performance for three images and two videos \textbf{maritime} datasets. Unlike generic results, we did not limit ourselves to objects with specific size and reported the results for the whole dataset, due to the fact that most of the objects are small. ``{*}" indicates the results only on the visible range videos.}
	\label{tab9}
	\centering
	\scalebox{0.88}{\begin{tabular}{@{}|ll|ll|ll|ll|@{}}
\toprule
\multicolumn{2}{|l|}{\multirow{2}{*}{Method}}          & \multicolumn{2}{l|}{SeaDronesSees}          & \multicolumn{2}{l|}{WSODD}          & \multicolumn{2}{l|}{ShipRSImageNet}        \\ \cmidrule(l){3-8} 
\multicolumn{2}{|l|}{}                                 & \multicolumn{1}{l|}{$\text{mAP}^{@0.5}$ $\uparrow$ } & $\text{mAP}^{@[0.5,0.95]}$ $\uparrow$  & \multicolumn{1}{l|}{$\text{mAP}^{@0.5}$ $\uparrow$}  & FPS $\uparrow$ & \multicolumn{1}{l|}{$\text{mAP}^{@[0.5,0.95]}$ $\uparrow$} & $\text{mAR}^{@[0.5,0.95]}$ $\uparrow$\\ \midrule

\multicolumn{1}{|l|}{\multirow{3}{*}{Image}} & SSD~(ECCV16)\cite{liu2016ssd} &  \multicolumn{1}{l|}{--}         &   --     & \multicolumn{1}{l|}{41.5}         &   43.02     & \multicolumn{1}{l|}{48.3}         &    61.8          \\ 

\multicolumn{1}{|l|}{} & Faster R-CNN+FPN~(NIPS15)\cite{ren2015faster} &  \multicolumn{1}{l|}{30.1}         &    14.2     & \multicolumn{1}{l|}{32.3}         &      19.42    & \multicolumn{1}{l|}{54.3}         &      --      \\ 

\multicolumn{1}{|l|}{}                       & Faster R-CNN+FPN~(CVPR17) \cite{xie2017aggregated} & \multicolumn{1}{l|}{54.7}         &   30.4       & \multicolumn{1}{l|}{--}         &   --      & \multicolumn{1}{l|}{--}         &     --     \\

\multicolumn{1}{|l|}{} & Mask R-CNN~(ICCV17)\cite{he2017mask} &  \multicolumn{1}{l|}{--}         &    --     & \multicolumn{1}{l|}{--}         &      --    & \multicolumn{1}{l|}{56.4}         &      --     \\

\multicolumn{1}{|l|}{} & RetinaNet+FPN~(ICCV17)\cite{lin2017focal} &  \multicolumn{1}{l|}{--}         &    --     & \multicolumn{1}{l|}{--}         &      --    & \multicolumn{1}{l|}{48.3}         &      68.9      \\ 

\multicolumn{1}{|l|}{}  &  YOLOv3~(arXiv18)\cite{redmon2018yolov3} &  \multicolumn{1}{l|}{--}         &    --     & \multicolumn{1}{l|}{56.1}         &   45.34       & \multicolumn{1}{l|}{--}         &      --     \\

\multicolumn{1}{|l|}{}  & TridentNet~(ICCV19)\cite{li2019scale} &  \multicolumn{1}{l|}{--}         &   --     & \multicolumn{1}{l|}{62.2}         &     10.16     & \multicolumn{1}{l|}{--}         &    --      \\

\multicolumn{1}{|l|}{}  & CenterNet-Hourglass~(arXiv19)\cite{zhou2019objects} &  \multicolumn{1}{l|}{50.3}         &    25.6     & \multicolumn{1}{l|}{--}         &    --      & \multicolumn{1}{l|}{--}         &  --   \\ 

\multicolumn{1}{|l|}{}  & CenterNet-ResNet~(arXiv19)\cite{zhou2019objects} &  \multicolumn{1}{l|}{36.4}         &    15.1     & \multicolumn{1}{l|}{--}         &     --     & \multicolumn{1}{l|}{--}         &     --     \\ 

\multicolumn{1}{|l|}{}  & CenterNet(ICCV19)\cite{duan2019centernet} &  \multicolumn{1}{l|}{--}         &    --     & \multicolumn{1}{l|}{53.5}         &     43.42     & \multicolumn{1}{l|}{--}         &      --     \\ 

\multicolumn{1}{|l|}{}  & FCOS+FPN(ICCV19)\cite{tian2019fcos} &  \multicolumn{1}{l|}{--}         &       --  & \multicolumn{1}{l|}{--}         &    --      & \multicolumn{1}{l|}{49.8}         &      67.4    \\

\multicolumn{1}{|l|}{}  & YOLOv4(arXiv20)\cite{bochkovskiy2020yolov4} &  \multicolumn{1}{l|}{--}         &       --  & \multicolumn{1}{l|}{57.2}         &      46.25    & \multicolumn{1}{l|}{--}         &      --     \\ 

\multicolumn{1}{|l|}{}  & FoveaBox(TIP20)\cite{kong2020foveabox} &  \multicolumn{1}{l|}{--}         &       --  & \multicolumn{1}{l|}{--}         &    --      & \multicolumn{1}{l|}{45.9}         &     62.2     \\

\multicolumn{1}{|l|}{}  & YOLOv3-2SMA(IJARS20)\cite{li2020modified} &  \multicolumn{1}{l|}{--}         &    --     & \multicolumn{1}{l|}{56.9}         &    \textbf{50.46 }     & \multicolumn{1}{l|}{--}         &      --     \\

\multicolumn{1}{|l|}{}  & EfficientDet-D0~(CVPR20)\cite{tan2020efficientdet} &  \multicolumn{1}{l|}{37.1}         &    20.8     & \multicolumn{1}{l|}{31.3}         &     30.83     & \multicolumn{1}{l|}{--}         &      --      \\

\multicolumn{1}{|l|}{}  & Cascade R-CNN~(TPAMI21)\cite{cai2018cascade} &  \multicolumn{1}{l|}{--}         &     --    & \multicolumn{1}{l|}{41.1}         &   29.56       & \multicolumn{1}{l|}{\textbf{59.3}}         &      \textbf{69.5}    \\

\multicolumn{1}{|l|}{}  & ShipYOLO(JAT21)\cite{han2021shipyolo} &  \multicolumn{1}{l|}{--}         &   --      & \multicolumn{1}{l|}{58.4}         &      49.81    & \multicolumn{1}{l|}{--}         &     --      \\ 

\multicolumn{1}{|l|}{}  & EfficientDet-D0+CroW~(ICCV21)\cite{varga2021tackling} &  \multicolumn{1}{l|}{--}         &    31.21     & \multicolumn{1}{l|}{--}         &    --      & \multicolumn{1}{l|}{--}         &    --        \\

\multicolumn{1}{|l|}{}  & YOLOv4+CroW~(ICCV21)\cite{varga2021tackling} &  \multicolumn{1}{l|}{--}         &    \textbf{36.41}     & \multicolumn{1}{l|}{--}         &     --     & \multicolumn{1}{l|}{--}         &      --           \\

\multicolumn{1}{|l|}{}  & Synth Pretrained RX101FPN~(arXiv21)\cite{kiefer2021leveraging} &  \multicolumn{1}{l|}{\textbf{59.2}}         &    32.6     & \multicolumn{1}{l|}{--}         &   --       & \multicolumn{1}{l|}{--}         &      --       \\

\multicolumn{1}{|l|}{}  & Synth Pretrained Yolo5~(arXiv21)\cite{kiefer2021leveraging} &  \multicolumn{1}{l|}{59.1}         &    33.2     & \multicolumn{1}{l|}{--}         &    --      & \multicolumn{1}{l|}{--}         &      --      \\

\multicolumn{1}{|l|}{}  & CRB-Net~(FN21)\cite{zhou2021image} &  \multicolumn{1}{l|}{--}         &      --  & \multicolumn{1}{l|}{\textbf{65}}         &    43.76      & \multicolumn{1}{l|}{--}         &    --      \\

\midrule
\multicolumn{2}{|l|}{\multirow{2}{*}{Method}}          & \multicolumn{1}{l|}{Seagull}          & \multicolumn{5}{l|}{SMD}                 \\ \cmidrule(l){3-8} 
\multicolumn{2}{|l|}{}                                 & \multicolumn{1}{l|}{ER $\downarrow$} & FPS $\uparrow$  & \multicolumn{1}{l|}{$\text{mAP}^{@0.3}$ $\uparrow$} & $\text{mAR}^{@0.3}$ $\uparrow$  & \multicolumn{1}{l|}{$\text{Pr}^{@0.5}$ $\uparrow$}   & $\text{Re}^{@0.5}$ $\uparrow$\\ \midrule

\multicolumn{1}{|l|}{\multirow{3}{*}{Video}} & ConvNet &  \multicolumn{1}{l|}{0.16}         &   --     & \multicolumn{1}{l|}{--}         &   --     & \multicolumn{1}{l|}{--}         &    --          \\

\multicolumn{1}{|l|}{} & Eigen-background~(TPAMI00) \cite{oliver2000bayesian} &  \multicolumn{1}{l|}{--}         &       --  & \multicolumn{1}{l|}{$0.5^*$}         &     $26.8^*$     & \multicolumn{1}{l|}{--}         &       --     \\ 

\multicolumn{1}{|l|}{} & Adaptive SOM~(TIP08) \cite{maddalena2008self} &  \multicolumn{1}{l|}{--}         &       --  & \multicolumn{1}{l|}{$1.2^*$}         &     $23^*$     & \multicolumn{1}{l|}{--}         &       --     \\ 

\multicolumn{1}{|l|}{} & Fuzzy ASOM~(NCA10) \cite{maddalena2010fuzzy} &  \multicolumn{1}{l|}{--}         &       --  & \multicolumn{1}{l|}{$1.5^*$}         &     $20.3^*$     & \multicolumn{1}{l|}{--}         &     --       \\

\multicolumn{1}{|l|}{} & LSTM &  \multicolumn{1}{l|}{0.22}         &         & \multicolumn{1}{l|}{--}         &      --    & \multicolumn{1}{l|}{--}         &       --     \\ 

\multicolumn{1}{|l|}{} & GRU &  \multicolumn{1}{l|}{0.17}         &    --     & \multicolumn{1}{l|}{--}         &    --      & \multicolumn{1}{l|}{--}         &    --        \\  

\multicolumn{1}{|l|}{} & GFLFM~(TCVPR15) \cite{xin2015background} &  \multicolumn{1}{l|}{--}         &       --  & \multicolumn{1}{l|}{$\textbf{8.9}^*$}         &    $ \textbf{32}^*$     & \multicolumn{1}{l|}{--}         &     --       \\ 

\multicolumn{1}{|l|}{} & Faster R-CNN~(NIPS15)\cite{ren2015faster} &  \multicolumn{1}{l|}{--}         &       -- & \multicolumn{1}{l|}{--}         &  --     & \multicolumn{1}{l|}{$81^*$}         &     $71^*$       \\ 

\multicolumn{1}{|l|}{} & YOLO~(CVPR16) \cite{redmon2016you} &  \multicolumn{1}{l|}{--}         &        -- & \multicolumn{1}{l|}{--}         &      --    & \multicolumn{1}{l|}{42.3}         &       57     \\ 

\multicolumn{1}{|l|}{} & SSD~(ECCV16)\cite{liu2016ssd} &  \multicolumn{1}{l|}{--}         &        -- & \multicolumn{1}{l|}{--}         &      --    & \multicolumn{1}{l|}{\textbf{83.7}}         &      40.1      \\ 

\multicolumn{1}{|l|}{} & Mask R-CNN recursive~(ICCV17)\cite{he2017mask} &  \multicolumn{1}{l|}{--}         &       --  & \multicolumn{1}{l|}{--}         &    --     & \multicolumn{1}{l|}{$78^*$}         &     $73^*$       \\ 

\multicolumn{1}{|l|}{} & Mask R-CNN fine-tuned~(ICCV17)\cite{he2017mask} &  \multicolumn{1}{l|}{--}         &       --  & \multicolumn{1}{l|}{--}         &    --     & \multicolumn{1}{l|}{$\textbf{82}^*$}         &    $71^*$        \\ 

\multicolumn{1}{|l|}{} & Mask R-CNN w/o seg.~(ICCV17)\cite{he2017mask} &  \multicolumn{1}{l|}{--}         &       -- & \multicolumn{1}{l|}{--}         &   --      & \multicolumn{1}{l|}{$\textbf{82}^*$}         &     $\textbf{77}^*$       \\

\multicolumn{1}{|l|}{} & Marie~\textit{et al.}(AVSS18) \cite{marie2018real} &  \multicolumn{1}{l|}{--}         &   --      & \multicolumn{1}{l|}{--}         &   --       & \multicolumn{1}{l|}{77}         &   79        \\

\multicolumn{1}{|l|}{} & ConvLSTM~(TGRS19) \cite{cruz2019learning} &  \multicolumn{1}{l|}{0.132}         &   --      & \multicolumn{1}{l|}{--}         &   --       & \multicolumn{1}{l|}{--}         &    --        \\ 

\multicolumn{1}{|l|}{} & ConvLSTM+DS Knowledge~(TGRS19) \cite{cruz2019learning} &  \multicolumn{1}{l|}{\textbf{0.13}}         &        -- & \multicolumn{1}{l|}{--}         &      --    & \multicolumn{1}{l|}{--}         &       --     \\ 

\multicolumn{1}{|l|}{} & CNN~(OSE20) \cite{leela2020image} &  \multicolumn{1}{l|}{--}         &        -- & \multicolumn{1}{l|}{--}         &      --    & \multicolumn{1}{l|}{--}         &      56     \\ 

\multicolumn{1}{|l|}{} & CNN+PASSTHROUGH L.~(OSE20) \cite{leela2020image} &  \multicolumn{1}{l|}{--}         &        -- & \multicolumn{1}{l|}{--}         &      --    & \multicolumn{1}{l|}{--}         &       68     \\ 

\multicolumn{1}{|l|}{} & CNN+PASSTHROUGH L. initialized~(OSE20) \cite{leela2020image} &  \multicolumn{1}{l|}{--}         &        -- & \multicolumn{1}{l|}{--}         &      --    & \multicolumn{1}{l|}{66}         &       73     \\ 

\multicolumn{1}{|l|}{} & Feng~\textit{et al.}~(TITS22) \cite{leela2020image} &  \multicolumn{1}{l|}{--}         &        -- & \multicolumn{1}{l|}{--}         &      --    & \multicolumn{1}{l|}{38.8}         &     \textbf{93.6}       \\ 
\bottomrule
\end{tabular}}
\end{table*}
\par\noindent\textbf{TinyPerson.} Table~\ref{tinyperson} shows the detection results obtained (\textit{i.e.,} MR and AP with IoU thresholds set to be 0.25, 0.5, 0.75) for the state-of-the-art methods on the images of tiny and small objects, whose number of pixels are in the range of [2,20] and [20,32], respectively. Recent methods are generally based on two commonly used object detection architectures, \textit{i.e.,} Faster RCNN-FPN and RetinaNet. Among these methods, MSM+~\cite{jiang2021sm+} achieved the best performance for almost all AP results. S-$\alpha$~\cite{gong2021effective} achieved the best results among the methods based on RetinaNet with respect to all MR evaluations. In contrast, MSM~\cite{yu2020scale} achieved relatively better results compared to other methods based on Faster RCNN-FPN in terms of all MR scores. Overall, the two-stage detection methods are seen to outperform the one-stage methods on TinyPerson.  
\par
\begin{figure*}
	
	\begin{tabular}{@{\hskip1pt}c@{\hskip3pt} c @{\hskip3pt} c @{\hskip3pt}}
		 SMD &\includegraphics[width=0.43\linewidth]{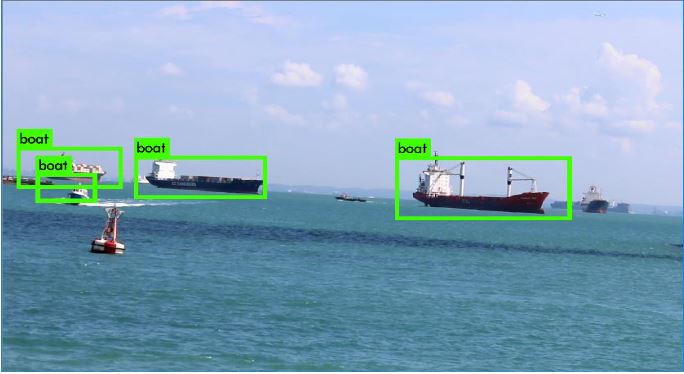} &\includegraphics[width=0.45\linewidth]{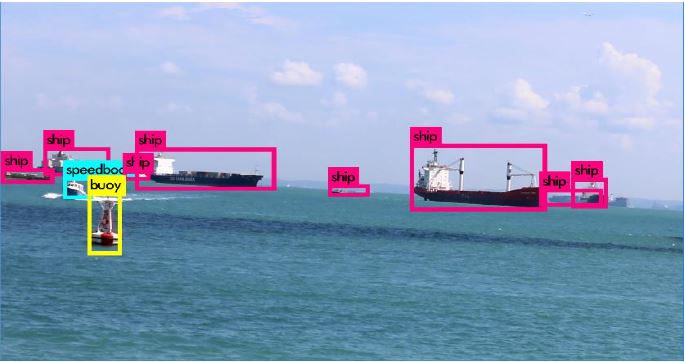}\\
		 &(a)&(b)\\
		\hline
	SeaShip &\multicolumn{2}{l}{\includegraphics[width=0.88\linewidth]{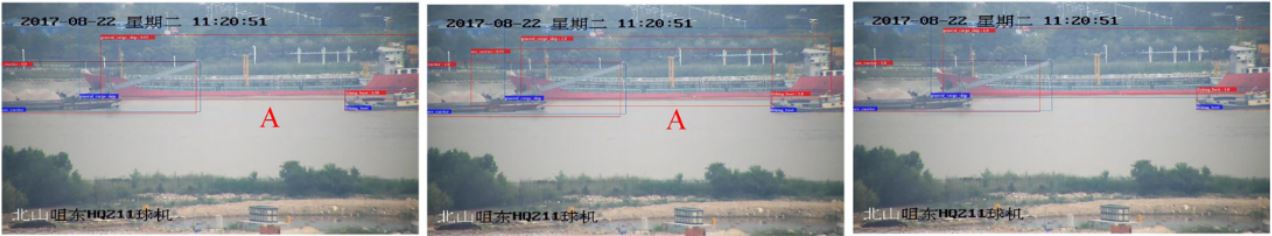}} \\
	 &(c)-(e)&\\
	\hline
	Seagull &\includegraphics[width=0.45\linewidth]{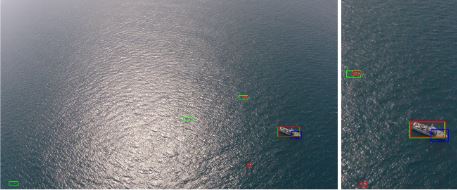} &\includegraphics[width=0.45\linewidth]{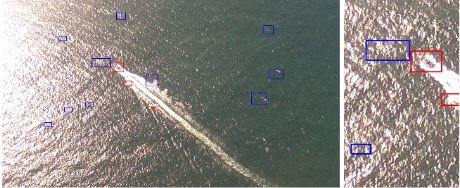}\\
		 &(f)&(g)\\
		\hline
	\end{tabular}	
	\caption{Examples of deep networks used for small object detection across several maritime datasets. (a) The baseline training was done on PASCAL VOC, and the test was done on SMD. (b) The result of training and testing the model on SMD. Image source: \cite{leela2020image}. (c)-(e) SeaShip dataset results. (c) SSD results,  (d) RefineDet, (e) results of \cite{liu2021attention} where blue box represents ground truth and red box represents predicted box. These images are from \cite{liu2021attention}. (f)\&(g) show the results on Seagull dataset. The green, blue, and red boxes are the outcomes of YOLO, detectnet+MHT and ConvLSTM. Images are from \cite{cruz2019learning}.  }
	\label{fig6}
\end{figure*}

\noindent\textbf{Other Maritime Image and Video Datasets.}
Table \ref{tab9} presents  detection results for other maritime datasets and the best results are marked in bold. The Table provides more information and identifies the leading methods for each metric. Figure \ref{fig6} shows some of the predicted bounding boxes for different datasets and techniques. Generally, it is observed that using general object detection frameworks to detect small objects is challenging, whereas small object specific methods can better locate those objects.
\section{Discussion and Future Directions}\label{discussion}

\subsection{Limitations}
Our review of the literature on the detection of small objects has identified several limitations, which are summarized in this section.
\begin{itemize}
\item Transformer models have recently greatly benefited computer vision and object detection in general, however the field of SOD has yet to fully utilize them. This is particularly more acute for video-based SOD.
\item While several studies have been conducted on generic SOD tasks, they either used different definitions of small objects, or they missed to report their experiments on publicly available datasets devoted to small objects, or they used a subset of a generic dataset with relatively large objects. Using MS COCO as an example:  (\textbf{i}) this dataset is not ideal for studying small objects;  (\textbf{ii}) different definitions are used for small objects (\textit{e.g.}, $32\times32$ or $16\times16$); or (\textbf{iii}) a small subset of small objects is used, which can result in bias and make benchmarking difficult. Due to these variations, comparing different techniques is generally difficult and challenging.
\item The technology of video-based small object detection (VSOD) is still evolving compared to image-based SOD, and only a few works use temporal information to detect objects 
\item There has not been any proper benchmarking of maritime SOD literature yet, and studies seldom use the same large-scale datasets. When it comes to VSOD, speed and the ability to monitor the maritime environment in real time are crucial. Recent studies overlook this and do not report FPS, which is vital for monitoring maritime environments in real time.
\item The majority of studies (mostly in maritime applications) apply popular models such as YOLO directly with only minor modifications, leading to poor performance of the given SOD application.
\item The mAP of large object detection techniques is generally high. On the other hand, the precision values of SOD methods are still low, requiring further investigation in the future.
\end{itemize} 
\subsection{Future Directions}
Taking into account the limitations of the reviewed works, we suggest the following directions for future research in SOD:
\begin{itemize}
    \item In light of the promising results achieved by transformer-based deep learning methods when applied to image-based generic small object detection, we believe that this model has the potential to achieve superior results in VSOD as well as SOD in maritime environments
    \item For a fair benchmarking, researchers should report their performance results on large-scale datasets such as {Tsinghua-Tencent 100K}, {CURE-TSD}, {USC-GRAD-STDdb}, {DOTA}, {VisDrone2021} for generic SOD and {TinyPerson}, {ETRI-Maritime}, {MOBDrone}, {Seagull}, {SMD}, {SeaDronesSees} for maritime SOD.
    \item The majority of current research exploits spatial information from videos and does not fully explore the temporal information; however, spatial and temporal information can be used together to minimize false alarms and miss detections for small objects when video quality is poor or when objects are occluded, which is especially relevant in maritime applications.
    \item The majority of prior studies have attempted to improve accuracy of SOD methods, but this has resulted in increased computational complexity, which is not desirable for real-time surveillance. Therefore, it is necessary to investigate networks that are accurate and lightweight.
    \item Even though multi-task or joint learning pipelines have yielded promising results for global feature extraction for small object identification, this area has not been studied deeply, and only a few papers have been published in this field
    \textcolor{black}{\item A majority of approaches reported in the SOD literature are based on the standard 2D-CNN. Hence, 3D-CNN can be used as an alternative to extend the 2D-CNN-based methods  for videos.  Moreover, the definition of small objects in images that deal with limited spatial information can be extended to video. In video, small objects can be redefined as objects with limited spatio-temporal information. Here, a limited temporal information refers to the fact that a small object (spatially small) appears in only a few frames of a video. With this new definition, all the existing tools for SOD using 2D-CNN can also be applied to 3D-CNN, such as pyramidal networks.  }
    \item In spite of the fact that most maritime objects are small (since the camera-to-object distance is large), analyzing the taxonomy of the works in the two domains (\textit{i.e.}, generic vs maritime), some ideas have been applied to only one domain whereas the other domain has not taken advantage of them. Following, we examine such ideas in both domains and discuss their potentials. (\textbf{i}) Although Super Resolution has improved generic SOD performance, it has not yet been investigated for maritime SOD. (\textbf{ii}) In maritime SOD, image enhancement is used to improve visibility under poor maritime conditions. It has not, however, been exploited for generic SOD. Then again, poor weather conditions may also hamper applications such as autonomous driving. (\textbf{iii}) Sea-Land Segmentation is another extensively used maritime SOD technique that reduces the number of false alarms. When prior information about the location of the objects is available, this approach could also be used for generic SOD. Pedestrians, for example, are not expected to appear in the sky. (\textbf{iv}) The use of context learning has been successful in improving generic SOD performance. Marine environments, however, do not lend themselves well to this method since water is a major component of the background.  (\textbf{v}) There have been limited studies examining the performance of recurrent networks for video-based detection, despite their success in sequential data analysis such as time series and natural language processing. 
\end{itemize}
\section{Conclusion}\label{conclusion}
In this paper, we survey more than 160 recent studies (2017-2022) in the field of small object detection in optical images and videos using deep learning, along with a maritime case study. A survey of  relevant  pre-processing techniques (\textit{e.g.}, data augmentation, super resolution), modern neural network architectures (\textit{e.g.}, 2D-CNN, 3D-CNN, RNN, transformers, and mixed architectures), feature learning (\textit{e.g.}, multi-scale, context, feature aggregation, and region proposal), multi-task learning, and loss function regularization for image and video-based small object detection is presented. In addition, 50 different datasets used for small object detection are extensively reviewed in this paper. This paper also presents popular learning and evaluation metrics and discusses their limitations. Lastly, potential future research directions in the field of small object detection are presented. 
\section*{Acknowledgement}\label{Acknowledgement}
This research is supported by the Commonwealth of Australia as represented by the Defence Science and Technology Group of the Department of Defence.
\ifCLASSOPTIONcaptionsoff
  \newpage
\fi
\bibliographystyle{IEEEtran}
\bibliography{reference}




\end{document}